\documentclass[12pt]{iopart}

\usepackage[utf8]{inputenc} 
\usepackage[T1]{fontenc}    
\usepackage{hyperref}       
\usepackage{url}            
\usepackage{booktabs}       
\usepackage{amsfonts}       
\usepackage{nicefrac}       
\usepackage{microtype}      

\usepackage[pdftex]{graphicx}
\usepackage{color}
\usepackage{siunitx}
\usepackage{listings}

\usepackage[square,sort,comma,numbers]{natbib}

\newcommand{\Reffig}[1]{Figure~\ref{#1}}
\renewcommand{\cite}[1]{\citep{#1}}

\newcommand{\n}[1]{\left\|#1\right\|}

\begin{document}
  
  \title[Data Augmentation for Brain-Computer Interfaces]{ 
  Data Augmentation for Brain-Computer Interfaces:\\
  Analysis on Event-Related Potentials Data
  }
  
 \author{
 Mario Michael Krell$^{1,2}$,
 Anett Seeland$^3$, and Su Kyoung Kim$^3$}
 \ead{\mailto{krell at uni minus bremen dot de}}
 \address{
 $^1$University of Bremen, Robotics Research Group}
 \address{
 $^2$International Computer Science Institute, Berkeley
 }
 \address{
 $^3$German Research Center for Artificial Intelligence, 
 DFKI GmbH, Robotics Innovation Center
 }

\begin{abstract}
  On image data, data augmentation is becoming less relevant due to
  the large amount of available training data 
  and regularization techniques.
  Common approaches are moving windows (cropping), scaling,
  affine distortions, random noise, and elastic deformations.
  For electroencephalographic data,
  the lack of sufficient training data is still a major issue.
  We suggest and evaluate different approaches to generate augmented data
  using temporal and spatial/rotational distortions.
  Our results on the perception of rare stimuli (P300 data) 
  and movement prediction (MRCP data)
  show that these approaches are feasible and can significantly
  increase the performance of
  signal processing chains for brain-computer interfaces
  by $1\%$ to $6\%$.
\end{abstract}


\submitto{\JNE}
\maketitle

\section{Introduction}
\label{sec:intro}

Brain-computer interfaces (BCIs) are becoming increasingly popular
with applications outside of the lab being pushed by companies like
Kernel, Facebook's Building 8, Neuralink, and Neurable.
BCIs detect a specific brain activity (e.g., in the electroencephalogram (EEG)),
which is correlated with the users intent (e.g., attention, movement preparation, etc.), 
and use it to  link a user with external systems, e.g., for
typing words or inferring the user's intentions
(for review see~\cite{Wolpaw:ClinNeuro:2002, Millan:FrontNeuro:2010}).
The interest of companies comes from a great progress 
that has been achieved in the last decade in EEG-based BCI applications
by using machine learning techniques.
In particular, single-trial detection of 
event-related potentials (ERPs), which correlates cognitives processes,
allows to deliver the users intent to external systems per event in real time
\cite{hillyard1983electrophysiology,regan1989human,sur2009event}.
For a more wide spread of BCIs, 
especially embedded brain-reading is interesting.
It uses single trials from EEG signals to infer human's intentions in real time
and thus allows to adapt the human-machine (or computer)
interaction (HMI, HCI)\cite{Kirchner2013,Kirchner2014PhD}.

However, before having BCIs running on large scale,
hardware and software still have to be optimized.
EEG data processing chains are usually handcrafted and 
the optimization is very difficult for various reasons:
a) EEG data is very noisy and non-stationary,
b) there are large differences in data between different recording days 
with the same subject and different subjects, and 
c) real-world applications do not often allow to record 
a large number of labeled training samples in a reasonable recording time.
Whereas some classical paradigms in the controlled conditions produce a decent amount of training data,
real-world applications do not.
For example for stroke rehabilitation,
it is sometimes not easy to get a sufficient amount of training data, since 
it is hard for the patient to perform more than $30$ movements due to fatigue. 
In other cases, e.g., for P300 or error potential data acquisition, 
the relevant stimulus needs to be rare to avoid 
that the user/subject gets annoyed or that the brain pattern changes.


To overcome the problems of transferability and an insufficient amount of data,
several approaches have been applied 
like ensemble learning~\cite{MetzenEnsemble2011,Lotte2015}, 
transfer learning~\cite{KindermansNIPS2012,Kindermans2014,Lotte2015,Jayaram2016}, 
and online learning~\cite{Shenoy2006,McFarland2011,Woehrle2015}.
Here, our approach is to modify the existing data to increase its amount 
and to support the learning algorithm in learning data invariances (data augmentation).
This has been so far done mainly 
in the temporal domain~\cite{Blankertz2006,Kirchner2013,kirchner2013_biodevices,Lotte2015,um2017data,Schirrmeister2017}, e.g., by different data segmentation techniques 
or by modifying the covariance matrix~\cite{Lotte2015,kalunga2015data}
but neither for P300 data nor for the spatial dimension.
We consider the spatial data augmentation to account
for moving caps and different head shapes
whereas the temporal distortion could compensate for varying
ERP onsets. 
Recently, data augmentation has been successfully used for deep learning
\cite{Schirrmeister2017,um2017data}.
Concerning adding noise to the data, the exhaustive analysis by Um et al. showed 
that the performance might increase in a few cases
but that there is no real benefit in it.
Instead of segmenting their signal into smaller chunks 
or having sliding windows, Um et al. used a permutation of the signal segments
to generate new data. 
Another very successful approach was to multiply the signal from 
a randomly chosen sensor by
$-1$, because for their signal of interest, the sign was not relevant.
This kind of rotation is not related to the movement of the cap but to 
properties of the data.
Both approaches were not applicable to event-related potentials
because here the order of segments and their sign is crucial.

Another motivation for data augmentation comes from the increasing interest
in applying deep learning algorithms to this kind of data
and its success in image data processing where 
it is common to apply different distortions \cite{Simard2003},
scaling, or moving windows/pixel shifts to create additional data 
to also make the
data processing more robust/invariant to these transformations.
For the break through in image data processing with deep learning,
data augmentation was crucial to avoid ``substantial overfitting''~\cite{Krizhevsky2012}.
An overview on recent approaches in EEG data processing with deep learning
was published not long ago by Schirrmeister et al.~\cite{Schirrmeister2017}.
Only recently, the spatial dimension of EEG data has been
paid attention to 
by either training a layer for spatial filters~\cite{Schirrmeister2017}
or by mapping the electrodes to an image~\cite{Bashivan2016}
which increases the dimensionality problem even more.
To fully consider the spatial dimension,
mapping strategies from multi-sensor arrays can be used
to find a compact ordering of the electrodes and to enable
convolutional filters. These are very similar to
the Laplacian reference in EEG data processing~\cite{Ramoser2000} 
but weights are not
fixed but learned (see also the appendix, Section~\ref{s:3d})

In this paper, we propose several data augmentation approaches for EEG data,
i.e., temporal and spatial/rotational distortions 
(Section~\ref{sec:meth}).
We evaluate especially the spatial approaches by comparing 
different parameterizations
(Section~\ref{sec:eval}).
We provide a conclusion and outlook in Section~\ref{sec:conc}.
This paper largely extends a paper with preliminary results 
obtained for rotational distortions~\cite{Krell2017EMBC}.

\section{Methods for EEG Data Augmentation}
\label{sec:meth}

In this section, we give an overview on the structure of EEG data 
and then introduce our approaches for the spatial as well as
the temporal part of the signal.

\subsection{Structure of EEG Data}

EEG data can be seen as two-, three-, or four-dimensional data.
The temporal dimension/axis is the most important one,
because the data comes with a very high resolution
(e.g., $5$kHZ)
that is directly related to the current brain activity.
Usually a maximum of $100$\,Hz is needed for data analysis.

The other dimension(s) correspond to the spatial component of the data
(different sensors/electrodes). 
This dimension is usually handled as a linear list 
with arbitrary sorting (1D) because the spatial resolution is very poor
and does not directly correspond to the real positions of the sources in
the brain.
However, as regards content, it corresponds to the sensors on the head surface
(2D) with positions in the 3D space.
In most EEG-based BCI applications/processing cases, 
electrode positions and the underlying spatial relations are not considered. 
In fact, there are correlations in the data between neighboring electrodes
and hence electrode positions should be taken into consideration\footnote{For 
further discussion, 
we refer to the appendix, Section~\ref{s:3d}.}.




\subsection{Spatial distortions}
\label{spatialD}
For EEG data processing, spatial robustness is a major issue.
When an EEG cap slightly shifts during experiments over time,
it is not easy to find the original places of electrodes 
and to reset the current positions of electrodes 
to their original positions accordingly.
Hence, there can be differences in electrode positions
between a first and second recording session 
or even during the same recording day
(spatial shifts within session and between sessions).
Furthermore, individually different head shapes of subjects
can contribute to differences in electrode positions between subjects
(spatial shifts between subjects).

One way to handle this issue is to use different modeling approaches,
e.g., exact measuring and tracking of electrode positions or
transforming the data to a space that is invariant from the electrode positions.
An example for such a space is the source space,
where the signal is mapped to anatomical regions inside the brain. 
However, for an accurate mapping the individual 
anatomy of the head as well as electrode positions of every recording session
are needed. 
For real-world applications, these requirements may be too cumbersome.

\begin{figure}[!t]
  \begin{center}
\includegraphics[width=.40\textwidth]{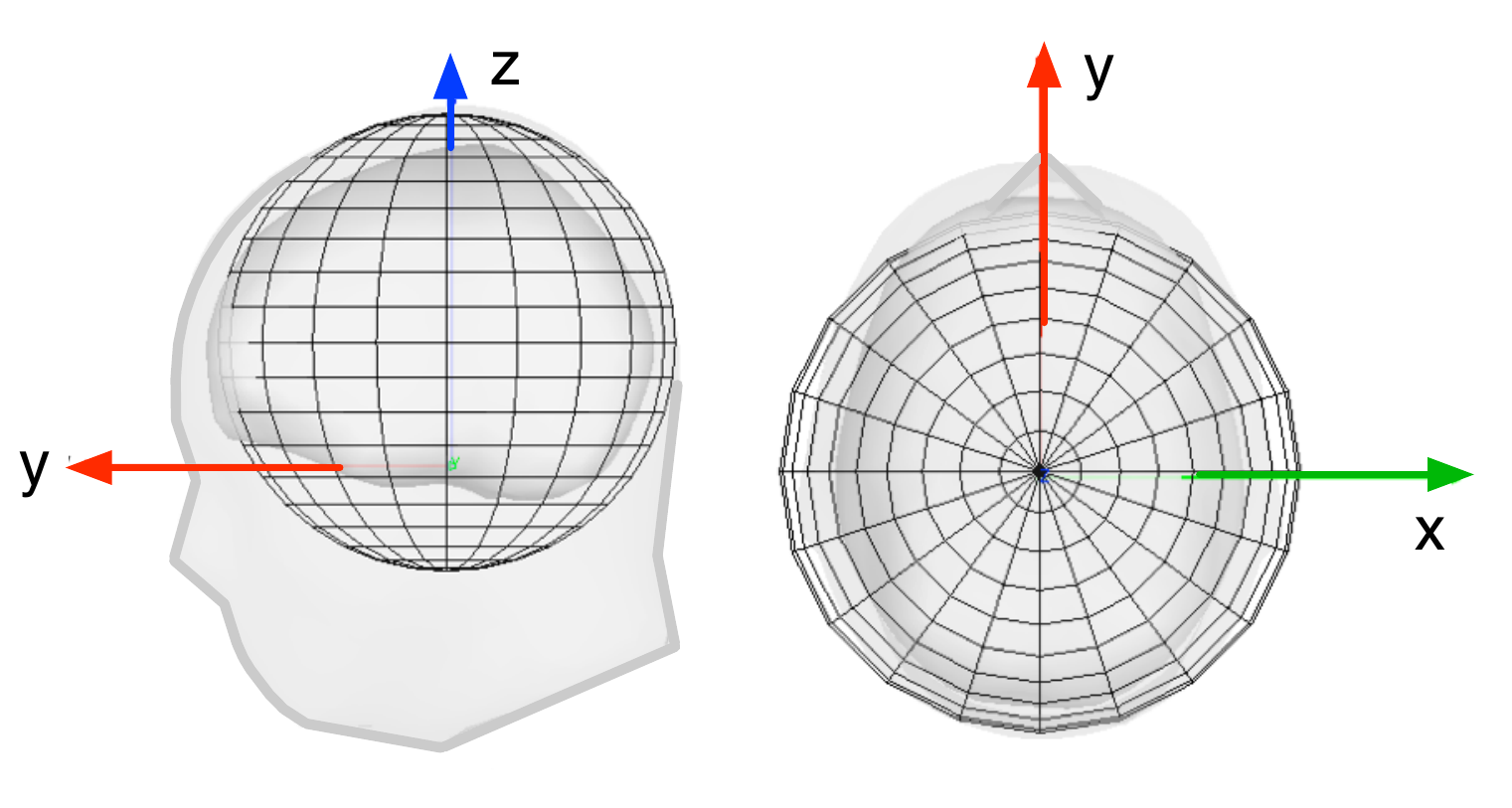}
 \includegraphics[width=.285\textwidth]{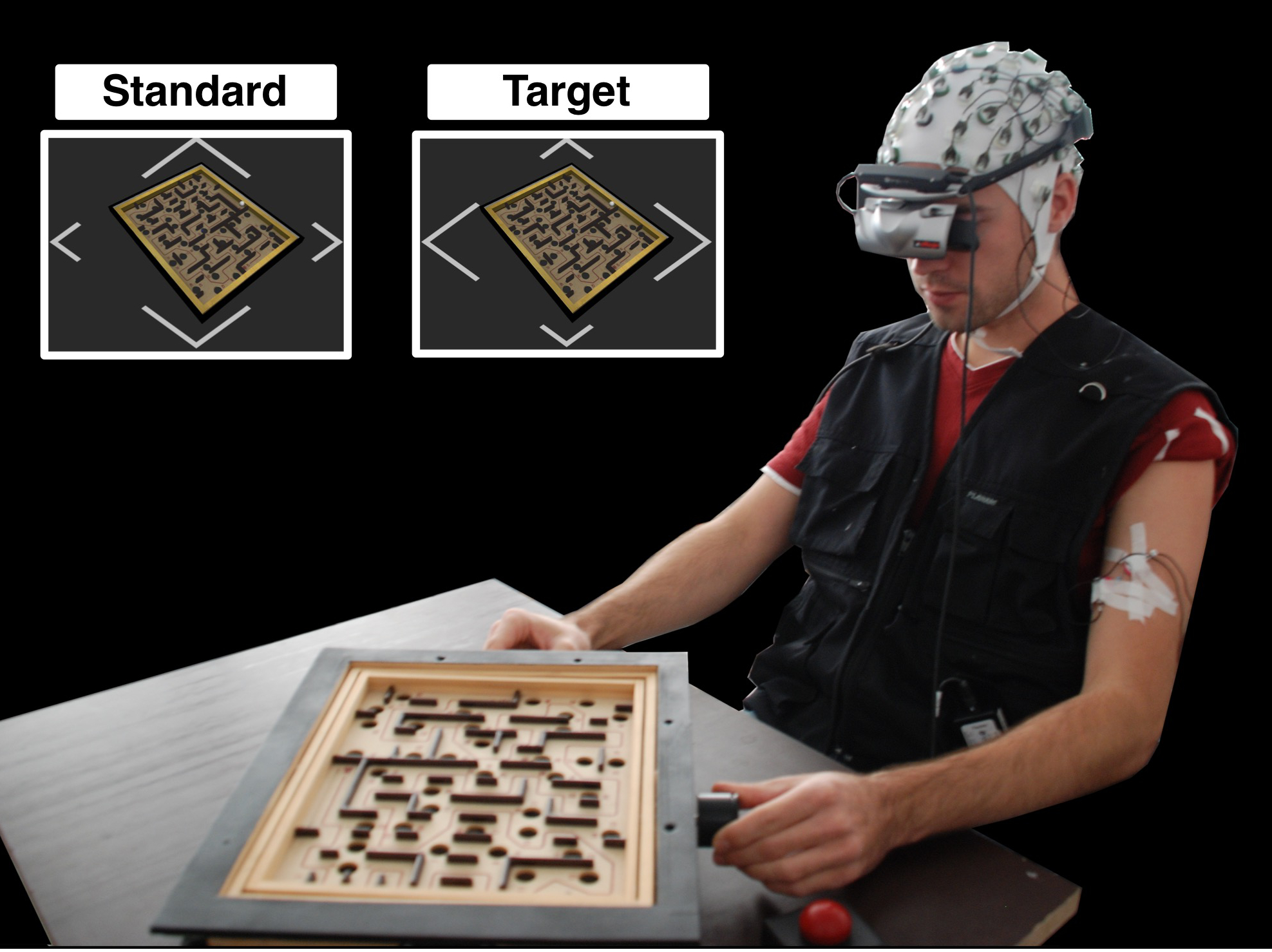}
     \includegraphics[width=.25\textwidth]{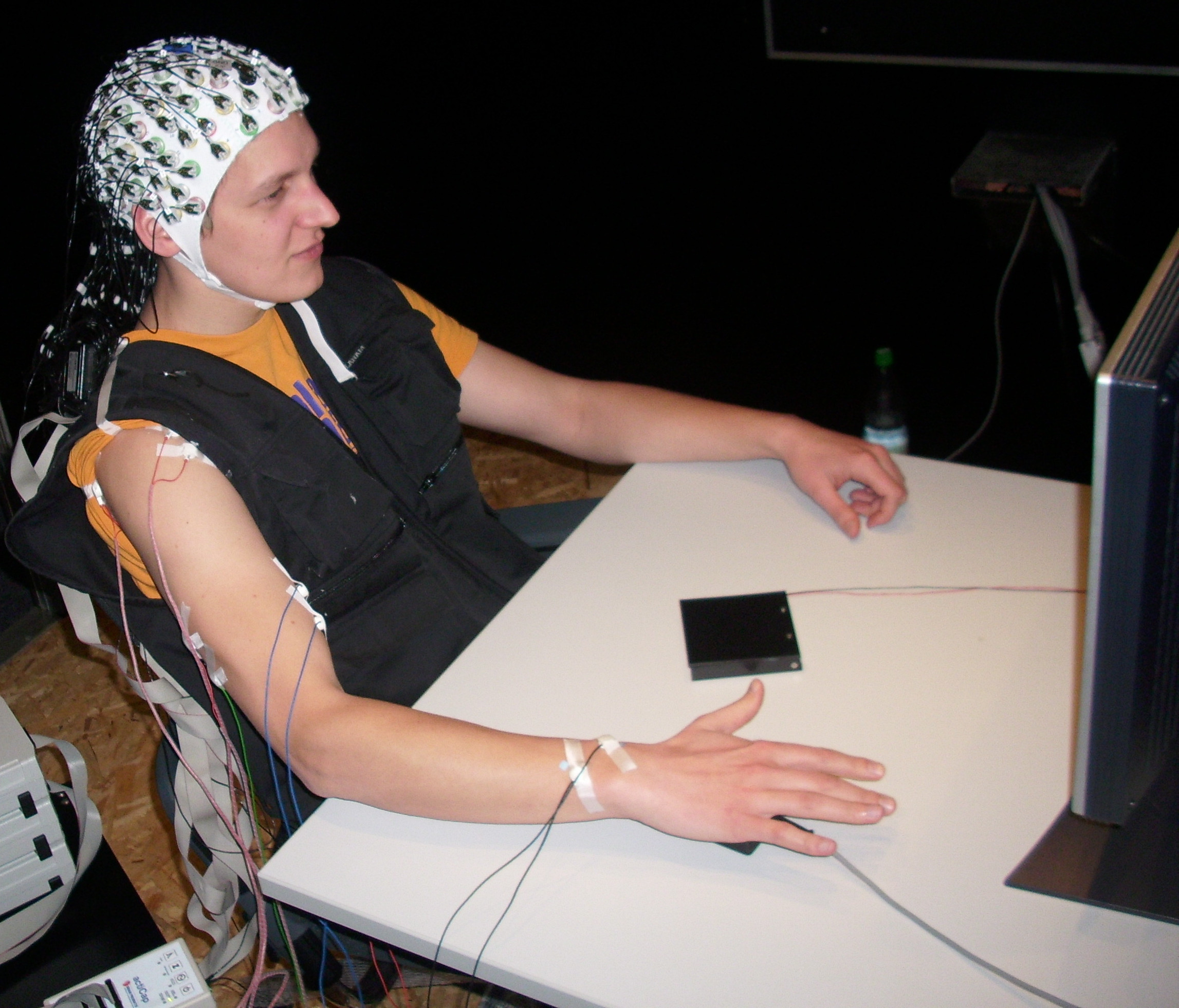}
  \end{center}
  \caption{[left] Spherical head model -- The x-axis points to the right, 
    the y-axis to the front, and the z-axis runs through the vertex.
    [middle] Recording setup P300 dataset -- The subject plays a virtual Labyrinth game 
    and reacts to the target stimuli with larger left and right corner
    by pressing the buzzer~\cite{Kirchner2013}. [right] 
    Recording setup MRCP
    dataset~\cite{Tabie2013} -- 
    The subject moves his/her hand voluntarily from a switch to a buzzer 
    while he/she sees a fixation cross on a monitor.}
    \label{fig:model}
\end{figure}

Our approach generates artificial data associated 
with differently shifted electrode positions
in a much easier way.
For reasons of simplicity, 
the way to generate new data is restricted 
to rotations around the three main axes
of the head as displayed in \Reffig{fig:model}[left].\footnote{This 
    graphic was created with Brainstorm \cite{Tadel2011}, 
    which is documented and freely available for download online 
    under the GNU general public license
    (\url{http://neuroimage.usc.edu/brainstorm}).
}
Since electrode positions and head-shape are usually unknown, we use
the standard positions, 
according to the extended $10$-$20$ system
extracted from analyzer2 and mapped to cartesian coordinates.
The new positions 
can be determined
by standard rotation matrices for the respective axes 
(e.g., 
$Q_x=[[1,0,0],[0,\cos(\theta),\sin(\theta)],[0,-\sin(\theta),\cos(\theta)]]$
for the x-axis).
So even though we are using the true positions in 3D,
we stay on the head surface (2D) by using only rotations and not
arbitrary distortions.
Furthermore, the relative positions of the sensors to each other are retained.
The data of the rotation is obtained by applying
the interpolation based on radial basis functions $\Phi$ (RBF)
like for example linear: $r$, cubic: $r^{3}$, and quintic: $r^{5}$ \cite{scipy}. 
For this interpolation, it is crucial that the data is normalized
beforehand such that electrodes show comparable ranges.
For each time point, the current amplitudes $f(p_i)$ from each sensor $p_i$
are taken 
and a new interpolation function $f$ is trained
to map electrode positions to the function values
\begin{equation}
  f(p_j)=\sum w_i \Phi(\n{p_j-p_i})\text{ with learned weights } w_i\, .
\end{equation}
Then the rotated coordinates are taken and the interpolation function
is applied to obtain the new function values (e.g., $f(Q_x p_i)$).
For significant speedup, the similarity to linear regression
and the fixed electrode coordinates in the interpolation can be used 
by storing the required matrix inversion (of the matrix $P^TP$ 
with 
$P = [\Phi(\n{p_i-p_j})]_{ij}$) at the beginning of the processing
and reuse is.

\subsection{Temporal distortions}
\label{sec:temp}
In most cases, EEG classification deals with
the detection of cognitive processes,
which are triggered by events that occur at specific time points.
For P300 classification, 
there are two different events:
a) rare occurring task-relevant events (\emph{targets}) and
b) more frequently occurring task-irrelevant events (\emph{non-targets}).
Around $300$\,ms after its presentation, 
the task-relevant event leads to a specific pattern 
in the brain called P300 \cite{Courchesne1977}.
The time point of presentation of an \emph{event} is marked in the EEG stream (event marker).
A common approach is to segment the continuous EEG based on these event
markers. 
Although the event markers are accurate, 
a latency delay of the P300 can occur for different reasons.
For example, it is possible that there is a delay between the display of the
events (on the monitor) and the perception of the events,
since the subjects have different visual focuses.
Workload changes during the experiment 
can also lead to a latency delay of the P300
\cite{Kok2001P3} and amplitude changes
(e.g., a reduction of the P300 amplitude \cite{Kirchner2016Frontiers}).
Note that the processing is usually resistant to different amplitudes (scaling)
of the pattern due to the aforementioned data standardization.

Another example of ERP classification is movement prediction based on 
movement-related cortical potentials (MRCPs) (e.g., 
\cite{bai_prediction_2011,seeland2013_ner,darmakusuma_analysis_2016}).
MRCPs include pre-movement components that mainly reflect 
preparatory processes of the brain before a movement \cite{shibasaki_what_2006}.
Since the preparation is a covert process where its onset in single-trial is unknown~\cite{sur2009event},
the \emph{event} that can be marked in the EEG is just the \emph{movement onset}.
Further, there is not even a perfect estimate
on when the movement starts, since usually the time point of the event has to be extracted 
from a different data stream and is not given by the experiment.
So using different data segments that are shifted
in time is promising here, too.

For data augmentation of EEG data, 
it might be also possible to stretch or compress the signal 
as it is common for image data.
This would be for example appropriate, 
when the pattern of an ERP is expected to change
in its length but not its latency.
Adding random noise to the data is not a good idea, because the data
already has a very low signal to noise ratio in contrast to image data.
If frequency changes were expected in the data,
the frequency domain of the time signal should be taken into consideration
and shifts or distortions should be performed in it~\cite{HuijuanYang2015}.
Also, if there is no real triggering event but a certain condition over time (e.g., imagining movements),
sliding windows could be used 
to cut out data segments from the data \cite{Schirrmeister2017}.

To keep the processing load for our evaluation low, we focus on 
shifting the marker of the triggering event.
This can be easily accomplished by cutting out additional segments according to
artificially shifted event markers (Details, see Section \ref{s:temporal_data_aug}).

\section{Evaluation}
\label{sec:eval}

In this section, both spatial and 
temporal data augmentation approaches were evaluated on the P300 and the MRCP data (Section \ref{s:data}).
In Section \ref{s:spatial_data_aug}, 
we analyzed different parameters and properties of the 
rotational distortion.
In Section \ref{s:temporal_data_aug}, 
the aspect of shifting the phase was evaluated.
All data processing was performed with the open-source
signal processing and classification environment
pySPACE~\cite{Krell2013} on a high performance cluster.\footnote{
The code for rotational distortions will become open-source
as part of pySPACE after acceptance.
}

\subsection{Dataset, Preprocessing, Classification}
\label{s:data}
For our experiments we used an initial simple setting on P300 data
and for the statistical evaluation, we analyzed the transfer
between subjects for P300 as well as for MRCP data.

The experimental scenario for P300 data 
(displayed in Figure~\ref{fig:model}[middle]) 
is based on an oddball paradigm
(for details see,
Kirchner et al. \cite{Kirchner2013} 
and the appendix, Section~\ref{sup:P300}).
Here, we used the recorded EEG data
from $5$ subjects.
Two recording sessions were collected per subject on two different days. 
Each session consisted of five runs and each run contained $720$ standards and $120$ targets.
We first analyzed the transfer between different recording days (inter-session)
on P300 data (sample size $10$)
and for statistical analysis we used the transfer between sessions
of different subjects (sample size $80$ due to more possible combinations).
For movement prediction based on MRCPs, data from $8$ subjects performing voluntary arm movements
with $3$ different speeds was used 
(for details see \cite{Tabie2013}, 
the appendix, Section~\ref{sup:MRCPdata}, 
and Figure~\ref{fig:model}[right]).
For every speed condition, 
three runs each containing $40$ movements were collected.
We merged all datasets of a subject and analyzed again the transfer between subjects. 
Both studies were approved by the ethic committee of the University of Bremen.

For processing, we used classical pipelines~\cite{Kirchner2013,seeland_biosignals2015}
consisting of stream segmentation, standardization, downsampling,
FFT frequency filter, xDAWN spatial filter, local straight line features,
a linear support vector machine (SVM)~\cite{Hsieh2008} 
with 2x5-fold cross-validation for
hyperparameter optimization,
and a threshold optimization to adapt the decision boundary 
to the chosen metric (details see the appendix, Sections~\ref{sup:P300proc} 
and \ref{sup:MRCPproc}). 
The rotational data augmentation was applied 
directly before the spatial filter.
To account for the unbalanced class ratio, we use the balanced accuracy
as performance metric, which is the arithmetic mean of true positive and
true negative rate \cite{Straube2014}.



\subsection{Rotational Data Augmentation}
\label{s:spatial_data_aug}
Before we evaluated our data augmentation approaches, 
we investigated the general effect of rotating the cap positions
on classification performance. Here, no data augmentation was applied 
(Section~\ref{s:cap_shif}).
We analyze the effect of different rotation axes (Section \ref{s:rot_axes}) 
and changes of data dimensions on classification performance 
(Section~\ref{s:dim_changes}).

Eventually, the core results are investigated with statistical verification
on subject transfer setups on the two different datasets.
For all cases, the effect of different rotation angles 
on classification performance was considered.


\subsubsection{Cap Shifts}
\label{s:cap_shif}

One idea underlying spatial robustness is based on 
the common knowledge:
a) The electrodes can be shifted during the recording session and 
b) The electrode positions will be usually 
different between recording sessions
(i.e., different recording days with the same subject).
The shift of electrode positions can be greater
between subjects.
Thus, the effect of shift of electrode positions on classification performance 
can be revealed within the same session and such effect
will be even stronger in the transfer between sessions and subjects.

\begin{figure}[!t]
  \begin{center}
\includegraphics[width=.49\textwidth]{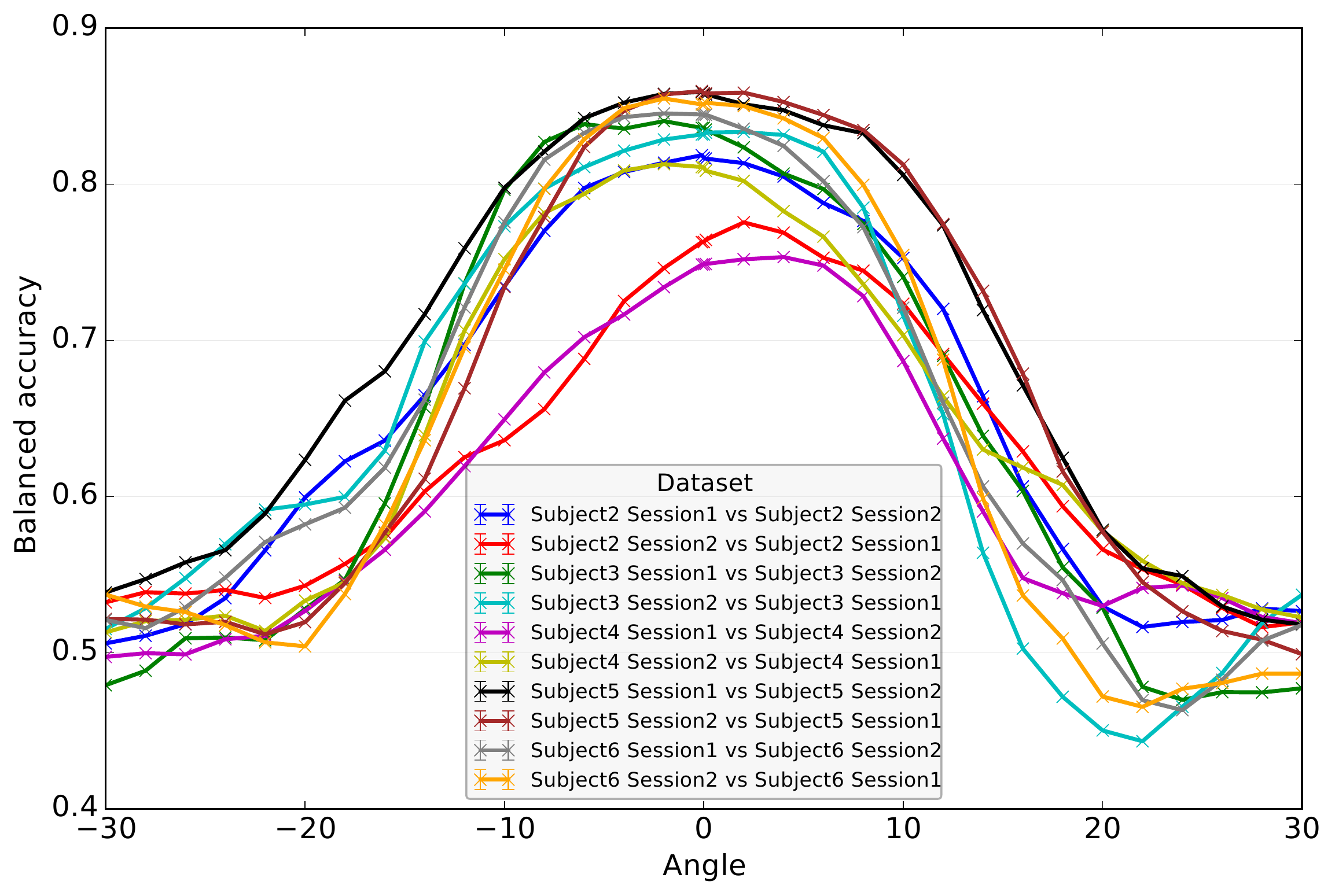}
\includegraphics[width=.49\textwidth]{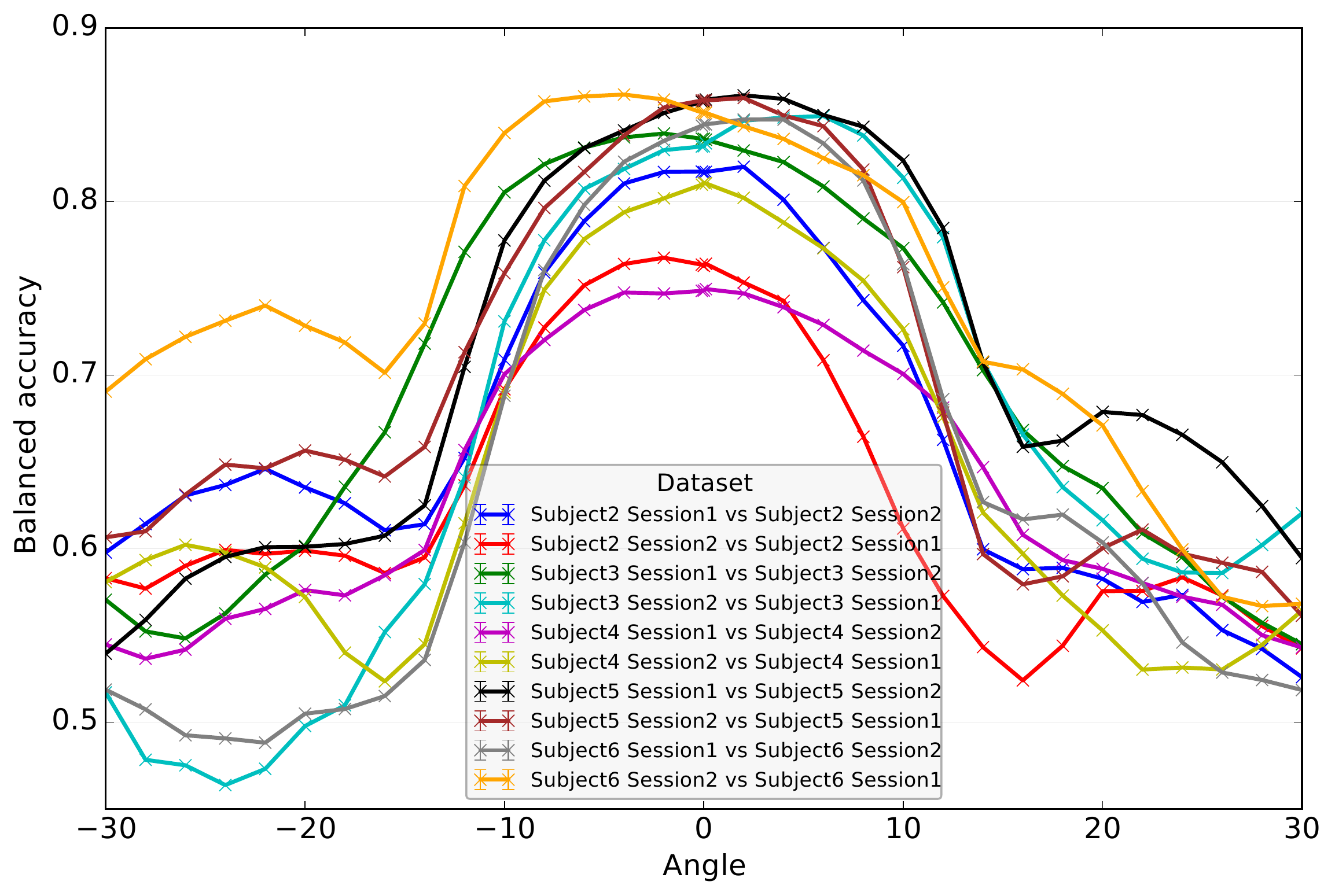}
  \end{center}
  \caption{Cap shift with different angles for a single axis
  (left: y-axis, right: z-axis) in the setting of transfer between sessions.
  No relevant effect for the x-axis could be observed.
  }
  \label{f:cap_shift}
\end{figure}

In this evaluation, 
we visualized the effect of shifted electrode positions and 
determine whether our interpolation method 
(details, see Section \ref{spatialD}) 
is capable of compensating for the shift of electrode positions.
Here, the data was not augmented but the original training data was replaced 
with a rotation around a single axis. The testing data was kept the same.
The expected shape would be a perfect bell curve, centered around zero.
For the different axes, the classification performances are depicted in \Reffig{f:cap_shift} for cap shifts with different angles 
except for the x-axis where no relevant effect could be observed.
For the y-axis, a deviation from optimum at an angle of zero 
can be observed in $7$ out of $10$ cases.
When the train and test data are exchanged,
the deviations from the zero angle should be reversely mirrored.
Such opposite pattern are apparent for subject 3 and subject 4.
For the z-axis, all subjects showed deviations and the opposite pattern except for one subject. Especially, the opposite pattern was 
obviously visible for subject 6.
All relevant angle shifts were between $-4\si{\degree}$ to $6\si{\degree}$.


Altogether, this indicates that cap shifts occurred between $-4\si{\degree}$ 
to $6\si{\degree}$ 
and can be (at least partially) compensated with our approach,
although other factors also seem to have an influence.
Furthermore, we can see that the influence is largest on the z-axis,
followed by the y-axis,
and it is probably irrelevant on the x-axis.
This effect was further analyzed in the next section
together with the data augmentation.

\subsubsection{Rotation Axes}
\label{s:rot_axes}

In this section, we analyze the effect of the three rotation axes
in the data augmentation for P300 data.
Here, we evaluate both single axis data and possible axes combinations.
For the data augmentation, we took the original data
and added an artificial sample for the chosen angle 
in positive and negative direction.

Using the x-axis reduces performance on average, 
whereas the augmentation around the y- and z-axis increases performance,
in which the z-axis slightly outperforms the y-axis
(for details we refer to the appendix, Section~\ref{sup:axex_inter}).
This is consistent with the observation
in Section~\ref{s:cap_shif} that the z-axis has the largest
effect of cap shifts and the x-axis is not affected by it.
The combination of y- and z-axis slightly increases classification performance,
whereas adding data, augmented with a rotation around the x-axis,
decreases classification performance in every combination.


The best performance was achieved with an angle between 
$12\si{\degree}$ and $24\si{\degree}$ over all subjects.
This pattern is different from the evaluation of 
a single axis, in which the much smaller angles were relevant
for the cap shifts (see Section~\ref{s:cap_shif}).
A possible reason is that the classifier interpolates for the 
angles in-between or generalizes better with 
this larger difference of the angles.

We also looked at the variability between subjects 
in the performance 
(see also the appendix, Section~\ref{sup:axex_inter})
Such subject-specific performance
should be considered, since we aim to 
configure the
data augmentation as independent from data properties as possible.
We observed a performance increase in $5$ out of $10$ cases.
In the other cases, there is no change or a slight decrease.
This absence of a substantial decrease 
is very important for the applicability of our approach.

\subsubsection{Data Reduction and Data Dimensionality Increase}
\label{s:dim_changes}
\begin{figure}[!t]
  \begin{center}
\includegraphics[width=.49\textwidth]{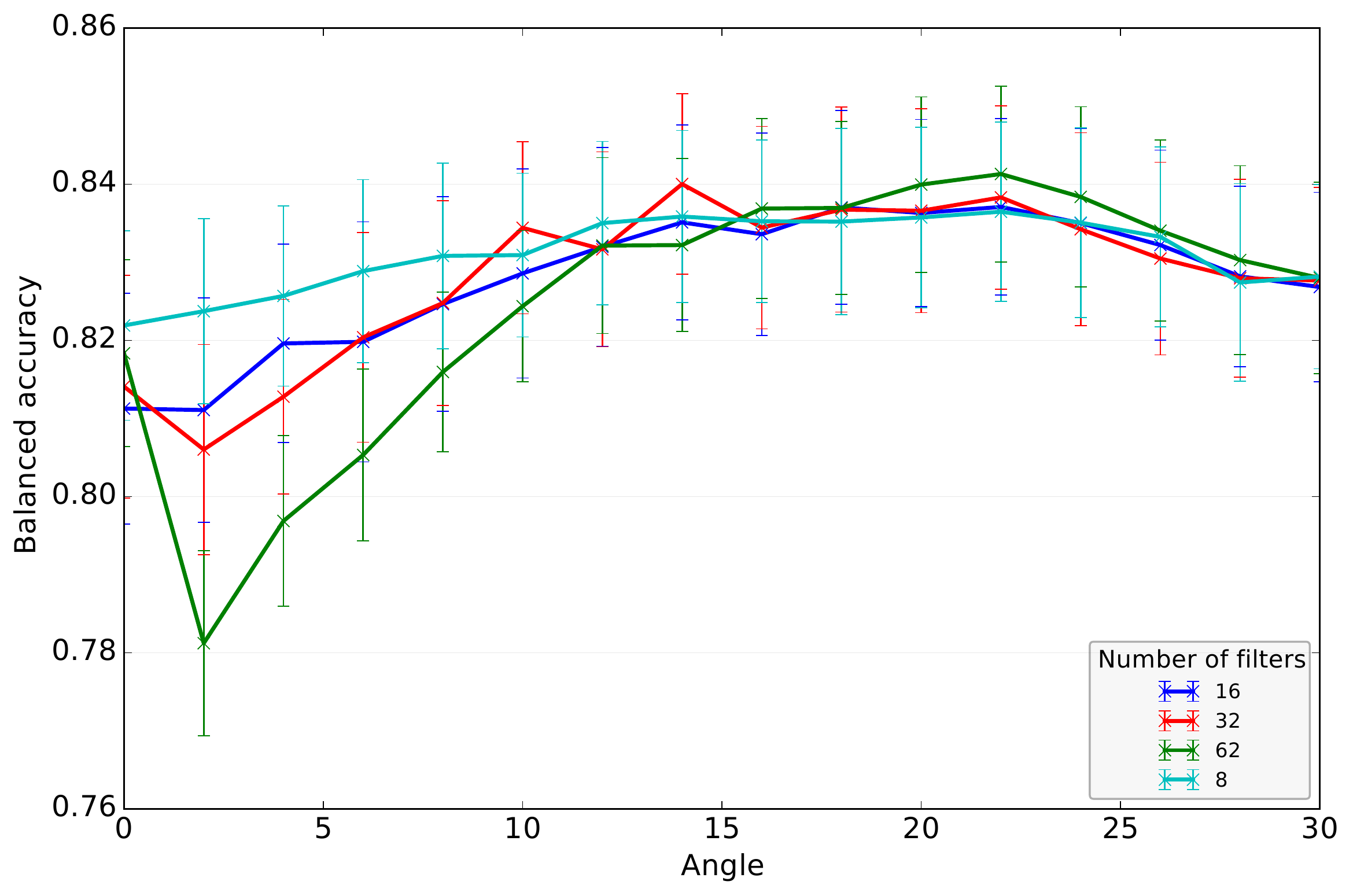} 
\includegraphics[width=.49\textwidth]{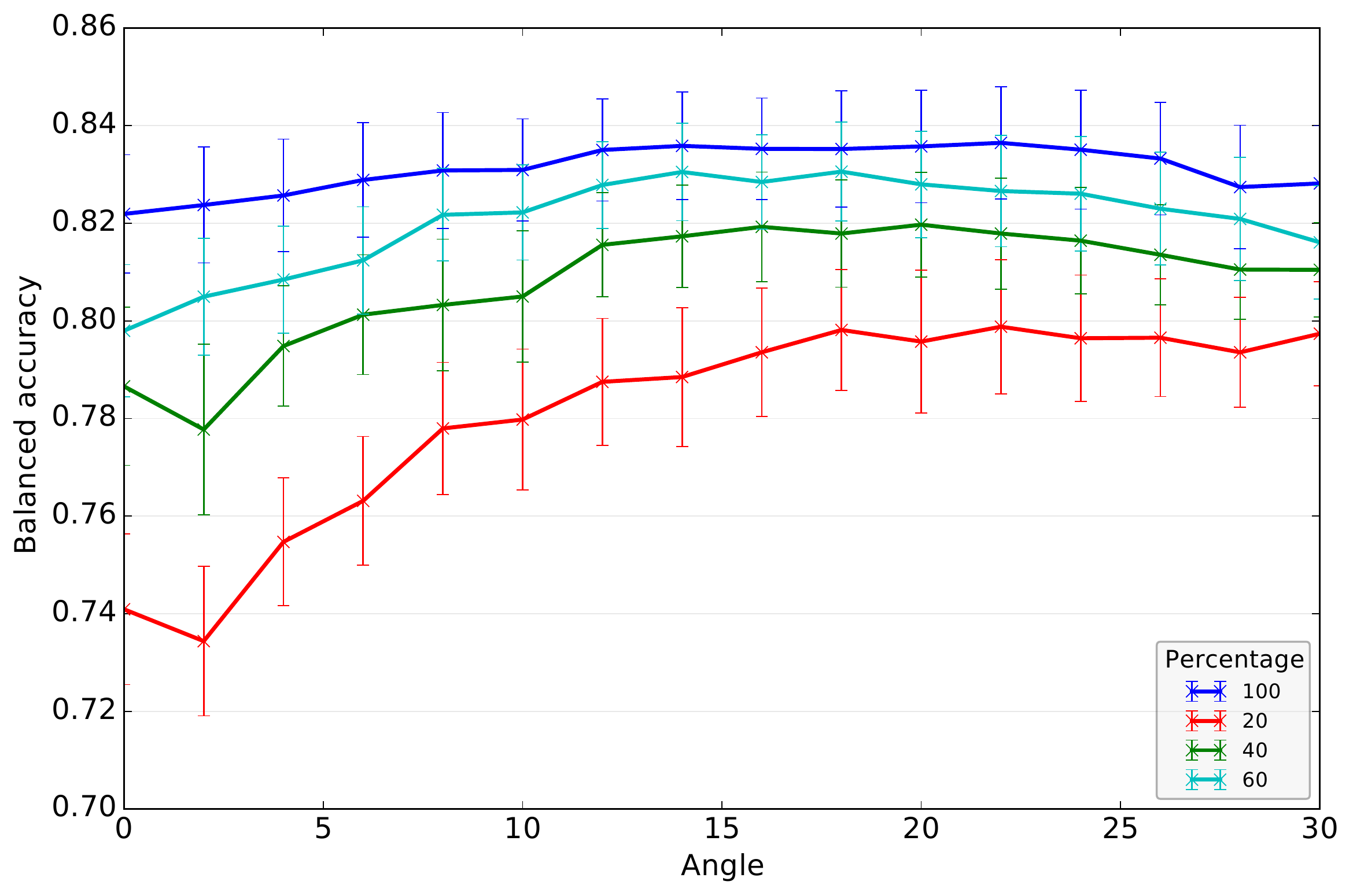}
  \end{center}
\caption{Comparison with different increased dimensionality 
(number of spatial filters -- left)
and data reduction
(percentage of used training data -- right).
}
\label{f:red}
\end{figure}

In general, the dimensionality correlates with the cap configuration (i.e., cap with different numbers of electrodes).
It is well known that machine learning algorithms behave differently depending
on the ratio between dimension of the data after the preprocessing and 
number of provided samples for each class.
For the xDAWN in the processing chain, 
the use of a larger number of filters increases data dimension 
for the SVM classifier.
For the case of using \emph{small rotation angles}, 
the performance was reduced due to an increased feature dimension
(see \Reffig{f:red} [left]). 
For $16$ filters, the dimension is doubled, and for $32$ it is quadrupled.
Fortunately, it is not relevant for larger angles over $14\si{\degree}$ (statistic details in the appendix, Section~\ref{sup:stat}).
This is very positive for our augmentation strategy because it
is still applicable when there is a lack of data.
The reason for the performance drop for small angles is probably that
the data augmentation is modeled by the classifier as noise and
degrades the classifier model whereas larger angles are modeled like new data.

\Reffig{f:red} [right] shows a similar effect when reducing the data size.
Again, there is a performance drop for small angels when $20\%$ or $40\%$ of
the data is used,
but still the performance is increased for rotation augmentation
between $18\si{\degree}$ and $20\si{\degree}$(statistic details in the appendix, Section~\ref{sup:stat}).
The overall performance is decreased due to the reduction of data size,
which is expected.
Note that the performance drop for small angles was not shown 
when sufficient amounts of data are available
($60\%$, $100\%$). 

\subsubsection{Subject Transfer}
\label{s:trans}
In this section, 
we investigated our rotational data augmentation approach
for the more challenging transfer setting 
between subjects with statistical verification.
In this transfer setting, 
we obtained $80$ performance values (sample size of $80$):
for each of the $5$ subjects, there were $2$ sessions for training,
which each were combined with testing on the data of the 
other $4$ subjects with $2$ sessions
($5$ subjects $\cdot$ $2$ sessions  $\cdot$ $4$ subjects $\cdot$ $2$ sessions) for P300 data.
We obtained $56$ performance values (sample size of $56$) for MRCP data: 
$8$ training datasets from the $8$ subjects combined with each of the 
remaining $7$ datasets for testing.
For the comparison between different axes (x, y, z) 
and their combination (y, z) in different angles,
we performed a two-way repeated measures ANOVA with \emph{axis} 
and \emph{angel} as within-subjects factors.
Note that we only combined y and z 
based on the previous evaluation on the session-transfer setting.
For the comparison between interpolation strategies in different angles,
we performed a two-way repeated measures ANOVA with 
\emph{interpolation strategy} and \emph{angel} as within-subjects factors.
For multiple comparisons, Bonferroni-Holm was applied.
Hence, we correct for multiple testing on the same dataset.

\begin{figure}[!t]
  \begin{center}
\includegraphics[width=.49\linewidth]{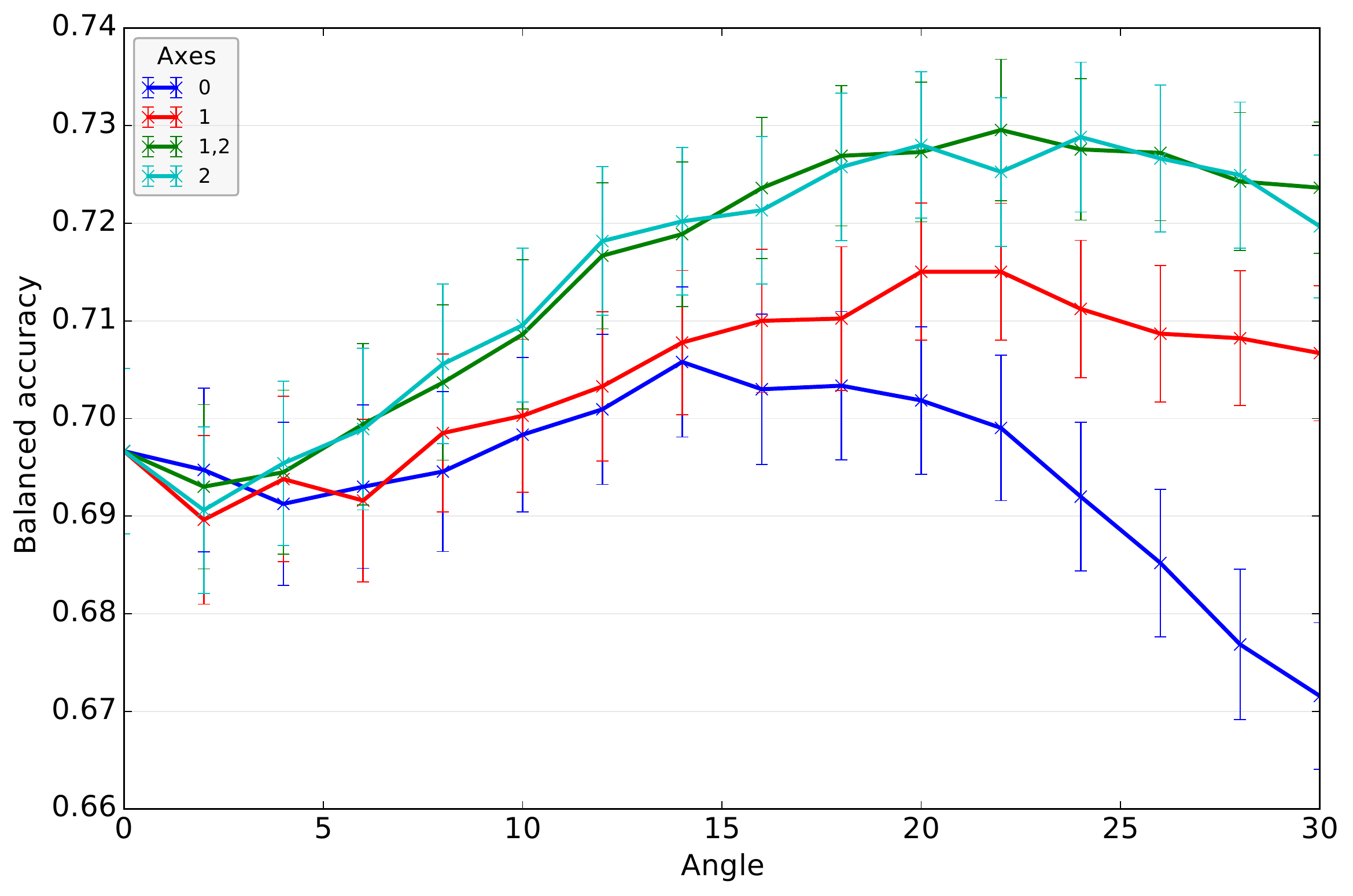} 
\includegraphics[width=.49\linewidth]{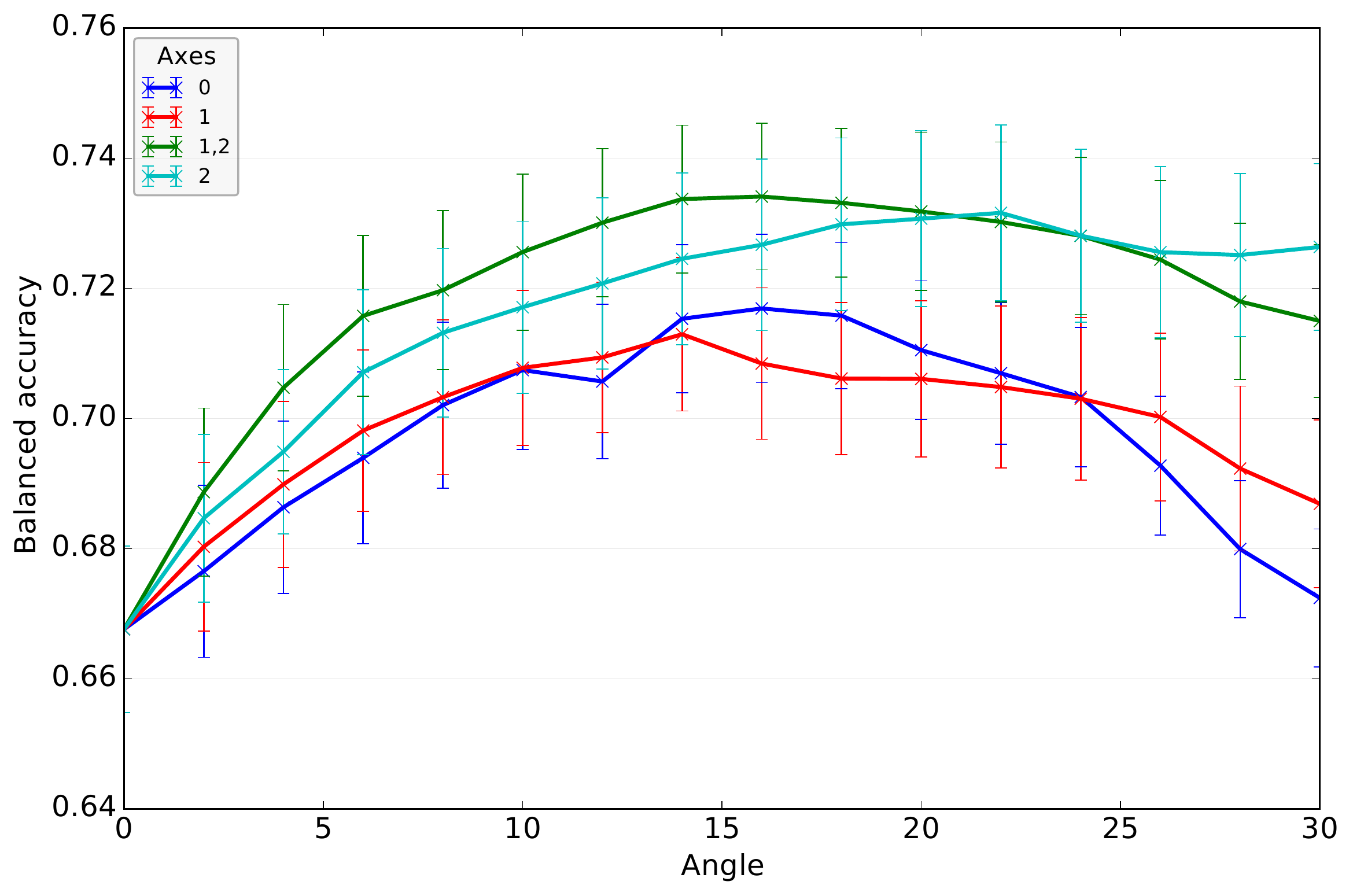} 
  \end{center}
\caption{Data augmentation with a rotation 
around the different axes: x~(0), y~(1), z~(2).
For the training the original data is taken 
and the augmented data with a rotation
angle of \emph{+angle} \textbf{and} \emph{-angle} is added.
This triples the amount of data and in case of 
combining the augmented data  by rotating around y and z~(1,2) it
is $5$ times the original data.
An angle of $0$ corresponds to the baseline with no augmentation.
See also \Reffig{fig:model} for the meaning of the axes.
For testing no data augmentation was applied.
[left] Subject transfer P300 data.
[right] Subject transfer MRCP data. 
}
\label{fig:sub_comp_ax}
\end{figure}

\begin{figure}[!t]
  \begin{center}
\includegraphics[width=.49\linewidth]{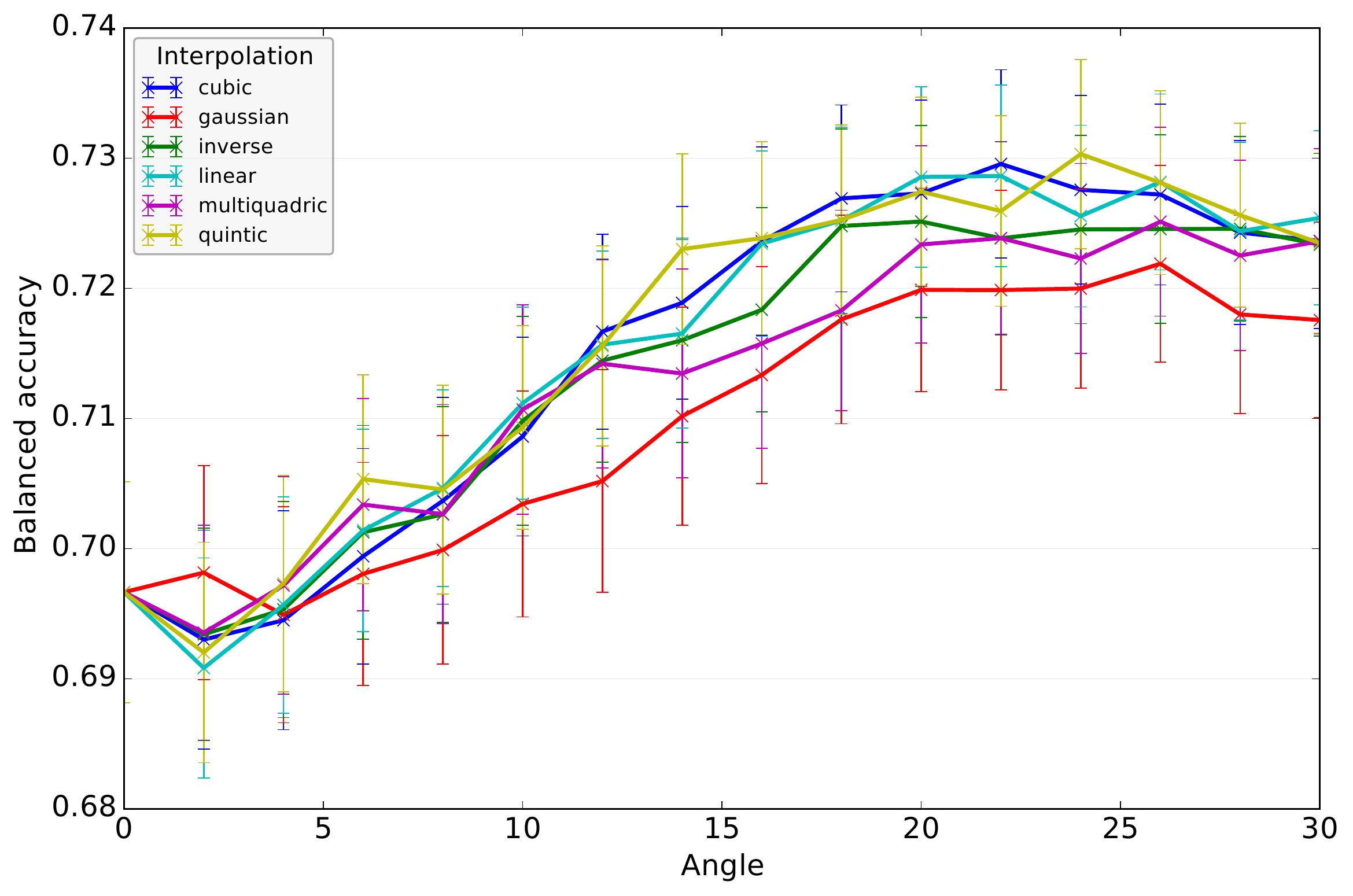}
\includegraphics[width=.49\linewidth]{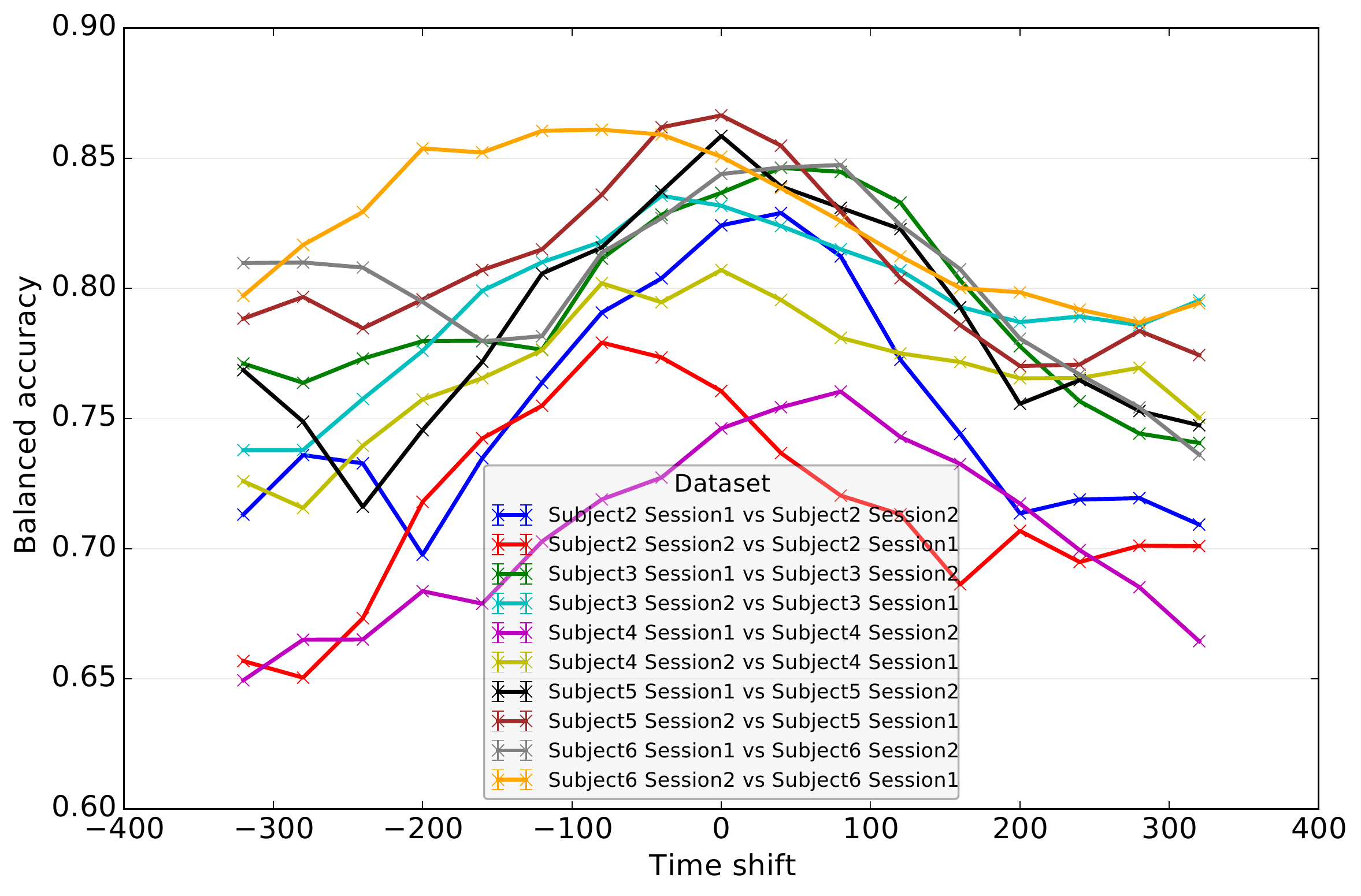}
  \end{center}
  \caption{
[left] Data augmentation with
rotation around the z- and the y-axis 
for different interpolation parameters (P300 subject transfer)
where $0$ corresponds to no augmentation.
[right] Comparison of different time shifts (in milliseconds) 
 (P300 session transfer).}
  \label{fig:time_shift_interpol}
\end{figure}

In \Reffig{fig:sub_comp_ax}[left], 
we compared different axes (with the ``cubic'' parameter
for the interpolation function according to the previous evaluation
in the appendix, Section~\ref{sup:inter_stra}).
The performance is clearly affected by the axes 
[$F_{3,237}$ = $23.9$, $p < 0.001$]. 
Again, the z-axis has the largest impact on the performance across angles
[$z$ vs. $y$: $p < 0.003$, $z$ vs. $x$: $p < 0.002$] 
but this time the combination with the y-axis does not further improve the performance over all angles
[$z$ vs. $z,y$: $p = n.s.$].
The overall performance is reduced due to the more difficult transfer setting 
(transfer between subjects) but rotations with an angle between 
$18\si{\degree}$ and $26\si{\degree}$
shows an increase in performance as beforehand 
[rotation effect: $F_{15,1185}$ = $5.50$, $p < 0.001$,
interaction effect: $F_{45,3555}$ = $1.07$, $p = n.s.$,
no rotation ($0\si{\degree}$) vs. between $18\si{\degree}$ and $26\si{\degree}$:
$p<0.05$, between $2\si{\degree}$ and $4\si{\degree}$ vs. between $18\si{\degree}$ and $26\si{\degree}$: $p<0.05$]
for z-axis and axis combination of z and y.
In \Reffig{fig:time_shift_interpol}[left],
we compared different interpolation strategies while rotating around the y- and z-axes.
All interpolation methods have a similar performance [interpolation effect: $F_{5,395}$ = $5.50$, $p = n.s.$,].
Again, the performance increases with rotations with an angle between 
$18\si{\degree}$ and $26\si{\degree}$ except for multiquadric and gaussian
[no rotation ($0\si{\degree}$) 
vs. between $18\si{\degree}$ and $26\si{\degree}$: $p < 0.05$]. 

We obtained a similar results on MRCP data as P300 data.
\Reffig{fig:sub_comp_ax}[right] shows 
an effect of the rotation angle on classification performance
[$F_{3, 165} = 15.02, p  < 0.001$].
Again, the z-axis achieves a better performance compared to x- and y-axis
[$z$ vs. $x$: $p < 0.001$, $z$ vs. $y$: $p < 0.002$]. 
Performance improvements through the axis combination is not observed as P300 data.
[$z$ vs. $z,y$: $p = n.s.$].
Further, improvements through rotations with an angle is observed both for z-axis 
and axis combination of z and y between $16$ and $30$ and between $10$ and $24$ respectively
[no rotation ($0\si{\degree}$) vs. between $16\si{\degree}$ and $30\si{\degree}$: $p < 0.03$ for \emph{z}-axis, 
no rotation ($0\si{\degree}$) vs. between $10\si{\degree}$ and $24\si{\degree}$: $p < 0.04$ for \emph{z,y}-axis].

\subsection{Time shift}
\label{s:temporal_data_aug}
In this section, 
we did not analyze spatial but temporal shifts.
The preprocessing was exactly the same but when cutting out segments,
different offsets were used to generate shifted data.
Similar to Section~\ref{s:cap_shif}, we first analyzed
the effect of the time shift on the classification performance.
Second, we augmented the data by adding data with positive
and/or negative offset to the data with zero offset.

For P300 data, using data with a non-zero offset slightly
improved performance for individual subjects
(\Reffig{fig:time_shift_interpol}[right])
which is reasonable due to the arguments in Section~\ref{sec:temp}.
Furthermore, a change in the sign of the optimal time shifts was
observed when comparing two sessions.
On average, performance decreased when the original zero offset was not used.

The results did not show a clear favoring time for augmentation
that holds for all subjects.
Hence, we tested the augmentation with
positive and negative time together.
Adding data with $40$ and $-40$ ms shift ($3$ times the training data) 
did not 
change performance on average but using larger offsets decreased performance
(see also the appendix, Section~\ref{sup:P300time}).
Hence, this approach does not add temporal robustness in our case 
(P300 detection) but at least increases the number of samples.

\begin{figure}[!t]
  \begin{center}
\includegraphics[width=.49\textwidth]{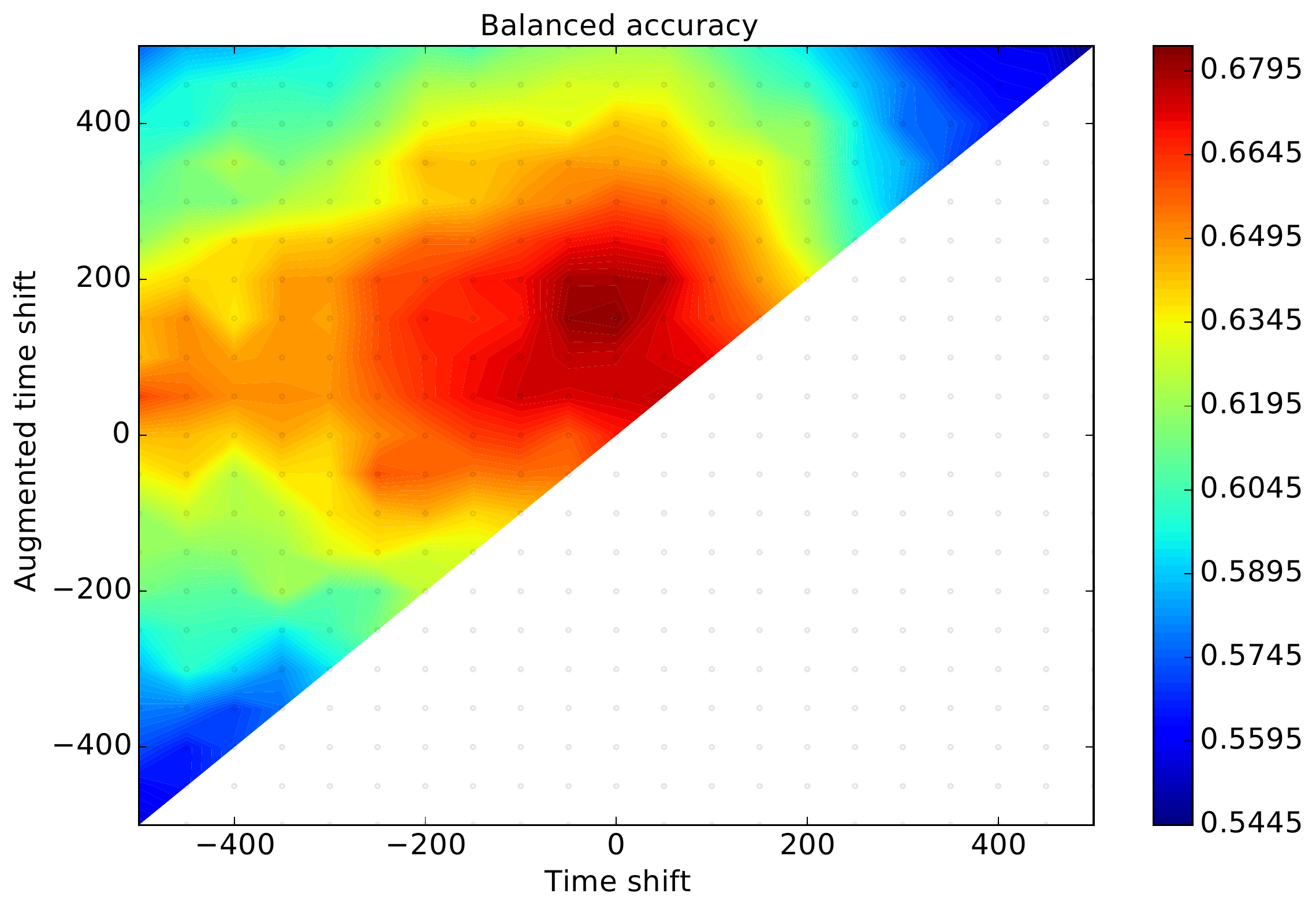} 
\includegraphics[width=.49\textwidth]{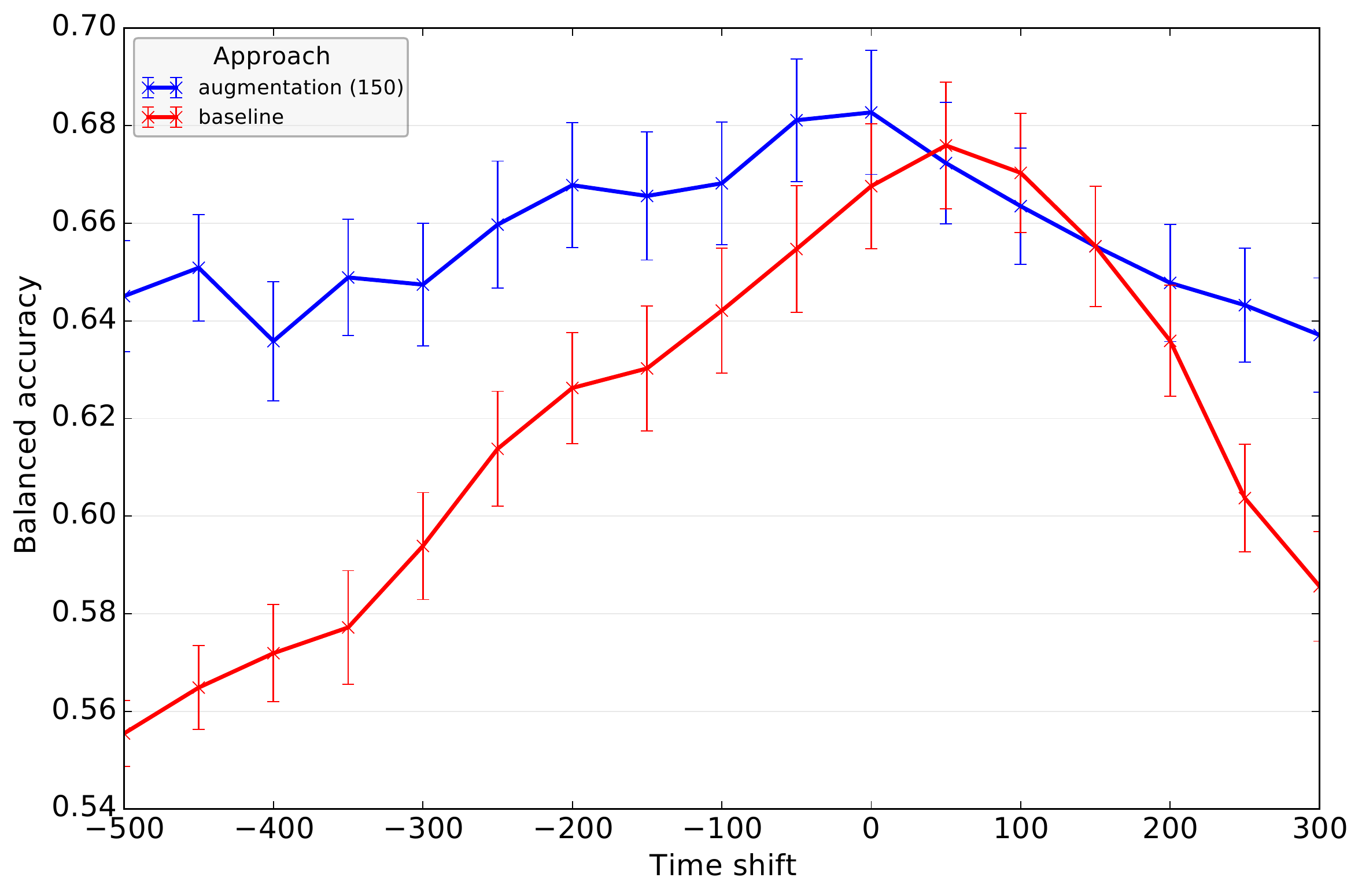} 
  \end{center}
\caption{Subject transfer MRCP data. [left] Classification performance of augmentation through combined time shifts (in milliseconds)
depending on an initial time shift. [right] Comparison of 
different single time shifts (red) and augmented time shift
of $150$\,ms (blue) depending on the initial time shift. Note, for the testing data neither initial nor augmented time shift was applied.}
\label{fig:lrp_time}
\end{figure}

For the investigation of the effect of augmentation through time shift
for the MRCP data,
we performed a two-way repeated measures ANOVA with
\emph{time shift} and \emph{augmentation} as within-subjects factors.
For multiple comparisons, Bonferroni-Holm was applied.
\Reffig{fig:lrp_time}[left] shows the classification performance 
when the training data used to detect movement preparation was augmented by a time shift.
The diagonal represents the \emph{baseline} where only a single time was used for training the classifier
(i.e., no augmentation equivalent to red line in \Reffig{fig:lrp_time}[right]).
The highest classification performance was not observed on the diagonal but with a time shift of $+150$\,ms, 
i.e., augmentation improves 
the performance (blue line in \Reffig{fig:lrp_time}[right]).
Obviously, both augmentation and time shift had an impact on the performance (\Reffig{fig:lrp_time}[right])
[augmentation effect: $F_{1,55} = 55.06, p < 0.001$, time shift effect: $F_{8,440} = 11.14, p < 0.001$]. 
Considering the results without an initial time shift (time shift $0$\,ms),
we observed performance improvement through data augmentation with a time shift of $150$\,ms
[baseline vs. augmentation: $p < 0.05$].
In comparison to the rotational augmentation, the increase in performance through time augmentation was smaller ($6$\,\% vs $1$\,\%). A possible explanation could be the subject transfer setting: best time shifts and best time shift combinations through augmentation might be subject specific. 



\section{Conclusion}
\label{sec:conc}

In this paper, we proposed several data augmentation techniques for EEG data
to generate new data without reducing the performance.
We analyzed and compared the behavior of these
on real EEG data for P300 detection and movement prediction (MRCP data).
Our analyses showed that 
the data augmentation 
did not substantially reduce the performance on average
with appropriately chosen configurations. 
In particular, concerning rotational shifts, we showed a significant 
increase of classification performance 
with a general setting 
of rotating only around the z-axis with an angle around $18$ degrees.
Additionally,
we obtained significant improvements on MRCP data with a temporal shift.

In future, we want to analyze further augmentation strategies,
use our findings for deep learning, and evaluate it on other paradigms.
Furthermore, it would be interesting to investigate whether complex head models or
aggregation of augmented results to one decision 
can improve the performance further.


\newcommand{\newblock}{}

\bibliographystyle{abbrv}
\bibliography{library,add_references}

\newpage
\appendix
\renewcommand\thefigure{\thesection.\arabic{figure}}  
\section{Structure of EEG Data and How to Obtain 2D-Representations of 3D Sensor Positions}
\label{s:3d}

EEG data can be seen as two-, three-, or four-dimensional data.
The temporal dimension/axis is the most important one,
because the data comes with a very high resolution of up to 
$5$kHZ that is 
directly related to the current brain activity.
Usually a maximum of $100$Hz is needed for data analysis.
Hence, it is common to standardize the data, reduce the frequency range
at least to less than $100$Hz and remove slow drifts
which can for example result from conductivity changes by sweating.

There are, of course, approaches that consider the 
spatial relations between electrodes
(e.g, source localization methods~\cite{michel_eeg_2004}, connectivity methods, etc.).
However, such methods require a considerable 
amount of expert knowledge and computational power
compared to the classical BCI applications.
Correlations between electrode measurements are considered 
by dimensionality reduction algorithms like spatial filters
without considering the real positions.

It is also possible to map the electrodes to a rectangular 2D-shape
to enable a spatial convolution instead or additionally to a spatial
filter layer in deep learning as outlined in the following.

For obtaining 2D coordinates that correspond to the true 3D positions,
there are three different methods.
Given the spherical coordinates $(r,\phi,\theta)$,
one possible mapping is 
\lstinline|x = r2 * numpy.cos(theta) * 60|, 
\lstinline|y = r2 * numpy.sin(theta) * 60| with
\lstinline|r2 = r / numpy.power(numpy.cos(phi), 0.2)|.
This representation is very common for plots of the signal but not that
useful for convolutional filter even though the resulting visualizations
could be used for applying standard image CNNs~\cite{Bashivan2016}.

But it is also possible to map the electrodes by hand to a rectangular shape
(see Table~\ref{sup:manual2D}).
Note, that this mapping is targeting a specific selection of sensors.
For a more general setting with arbitrary types of homogeneous sensors
an identification algorithm by Kampmann~\cite{Kampmann2016}(section 5.1.6) can be used.
First, a graph is constructed by the similarity between sensors 
(like distance or covariance) and rearranged by
the force-directed layout algorithm to have neighboring sensors 
to be also nearby in the 2D graph representation.
For the final mapping of the graph to a rectangular shape,
self-organizing map are used.

\begin{table}[ht!]
  \centering
  \caption{2D rectangular mapping of EEG electrode positions.}
\begin{tabular}{ccccccccc}
F7 & AF7 & FP1 & AF3 & Fz & AF4 & FP2 & AF8 & F8\\
FT9 & F5 & F3 & F1 & FCz & F2 & F4 & F6 & FT10\\
FT7 & FC5 & FC3 & FC1 & Cz & FC2 & FC4 & FC6 & FT8\\
T7 & C5 & C3 & C1 & CPz & C2 & C4 & C6 & T8\\
TP9 & CP5 & CP3 & CP1 & Pz & CP2 & CP4 & CP6 & TP10\\
P7 & P5 & P3 & P1 & POz & P2 & P4 & P6 & P8\\
PO9 & PO7 & O1 & PO3 & Oz & PO4 & O2 & PO8 & PO10
\end{tabular}
\label{sup:manual2D}
\end{table}

\section{P300 Data}
\label{sup:P300}
In the experimental scenario,
the subject saw a task-irrelevant event (standard) 
every second with a latency jitter of $\pm 100$\,ms.
With a probability of $1/6$, a task-relevant event (target) 
was randomly displayed. 
The task relevant event/stimulus
required a reaction from the subject (see also \Reffig{fig:model}[middle]).
The stimuli were chosen to be very similar to avoid the effect of color or shape on the brain pattern.
Based on this reaction to the targets,
we can infer the true label for standards and targets.
When the subject correctly responded to targets,
we can ensure that targets were correctly perceived.
The perceived task-relevant event 
leads to a specific pattern in the brain, called P300.
In this scenario, the continuous EEG was recorded from $5$ subjects
using a 64 channel actiCap system (Brain Products GmbH, Munich, Germany)
with reference at FCz (extended 10-20 system).
Two electrodes of the 64 channel system were used 
to record the electromyogram (EMG) which is related with to pressing of the buzzer.
Impedance was kept below $5$\,k$\Omega$.
EEG and EMG signals were sampled at $1$ kHz, amplified by two 32 channel BrainAmp DC amplifiers 
(Brain Products GmbH, Munich, Germany) and filtered with a low cut-off of $0.1$ Hz and high cut-off of $1$ kHz.
Two recording sessions were collected per subject on two different days. 
Each session consists of five runs and each run contains $720$ standards and $120$ targets.
Targets without a response by the subject were not considered in our evaluation.
The motivation of the previous study by Kirchner et al.~\cite{Kirchner2013} 
was to transfer the
trained classified to distinguish missed and perceived targets by
treating missed targets like standards due to similarities in the shape.
Hence, a few of the provided trials (missed targets) had to be removed
because no clear label could be assigned.
We first analyzed the transfer between different recording days (inter-session)
on P300 data
and for statistical analysis we used the transfer between sessions
of different subjects.

\section{MRCP Data}
\label{sup:MRCPdata}
Eight healthy, right-handed male subjects took part in the experiment
where they performed self-initiated movements.
Subjects sat in a comfortable chair in a dimly lit room, resting their arms
on a table in front of them.
Subjects placed their hand on a flat switch and saw
a fixation cross presented on a
monitor (see also \Reffig{fig:model}[right]).
A buzzer was placed 
approximately $20$\,cm to the right from the switch. 
Subjects could initiate whenever they want a movement from the switch to the buzzer
but a resting time of at least $5$\,s between consecutive movements had to be maintained.
Subjects were informed on the monitor if they violated that condition and the performed movement was marked as invalid.
Invalid trials were not considered in data analysis.
Each experimental run consisted of $40$ valid movements.
For each condition, e.g. different speeds, three runs were recorded.
 Subjects were first instructed
 to perform the movements in their normal speed.
 The subject performed fast and slow movements according to predefined speed restriction,
 which was individually calculated per subject.

EEG data was recorded with actiCAP ($128$ electrodes placed according to the extended 10-20 scheme)
and four BrainAmp DC amplifiers (all Brain Products GmbH, Munich, Germany).
FCz was used as reference and impedances were kept below $5$\,k$\Omega$.
Before storing the data with a sampling frequency of $5$\,kHz to disk, a band-pass filter between $0.1$ und $1000$\,Hz was applied.
The start and end of a movement was marked into the EEG stream based on events from the switch and buzzer, respectively. 

\section{P300 Preprocessing}
\label{sup:P300proc}

We segmented the continuous EEG based on each event 
with a segment length of one second and normalized them 
to zero mean and a standard deviation of one.
The sampling rate of the data was reduced from $1000$Hz to $25$Hz.
Then the data was low-pass filtered with a cut-off frequency of $4$Hz,
which was chosen based on our previous evaluations.

After this preprocessing step, the data augmentation approaches were applied
except of the temporal shifts. Those were already applied at the beginning 
of the processing chain during the cutting of
the segments.
Next, the xDAWN spatial filter \cite{Rivet2009}
was trained on the complete training data and then applied.
Afterwards, local straight lines were fitted 
and the respective slopes were taken as features.
Features were normalized to zero-mean and unit-variance on the training data.
The used classifier was a standard (affine) SVM implementation with a linear kernel
\cite{Hsieh2008} and limited number of iterations ($100$ times the number of samples).
The regularization hyperparameter $C$ of the SVM was optimized 
using $5$ fold cross validation with two repetitions and with the values 
$[10^{0},10^{-0.5},\ldots, 10^{-4}]$.
Eventually, a threshold optimization was applied.

For an \emph{inter-session} P300
evaluation, we train with the data of one recording session
and then test on the other remaining recording session of the same subject.
This results in $10$ samples ($5$ subjects * $2$ sessions).
Setting the cap anew at this other day, can result in slight
changes of the electrode positions.
For an \emph{inter-subject} P300
evaluation, we train with the data of one recording session
and then test on the recording sessions of the remaining subjects.
This results in $80$ samples ($5$ subjects * $2$ sessions * $4$ subjects * $2$ sessions).

\section{MRCP Preprocessing}
\label{sup:MRCPproc}
We segmented the continuous EEG based on the switch event (switch release).
Since event has some delay in detecting the actual movement onset,
we used a window of one second length ending $100$\,ms before the switch event.
In this way we could ensure that the upcoming movement is \emph{predicted} and not
just \emph{detected}. 
 For the opposite class, we cut out non-overlapping windows
 starting soonest $1$\,s after subjects entered the switch and ended latest $0.8$\,s before subjects left the switch.
 For the time augmentation we varied the end time of the windows for the \emph{movement preparation} class
 between $-0.6$ to $+0.5$\,s with respect to the switch release.
 
 First, the data was normalized channel-wise to zero mean and unit variance. 
 Next, a decimation from $5$\,kHz to $20$\,Hz was performed in two steps.
 Then, the data was band-pass filtered between $0.1$ and $4$\,Hz using a FFT.
 Subsequently the data was reduced to the last $200$\,ms since the most relevant
 information is expected in that range. At this point the rotational data augmentation was applied in our analysis.
 Afterwards the xDAWN spatial filter \cite{Rivet2009} was trained which reduced the
 dimension of the data to $4$ pseudo-channels. 
 These were directly used as features ($4$ channels $\times$  $4$ time points $= 16$ features),
 normalized (zero-mean, unit-variance) and passed to a standard (affine) SVM implementation with a linear kernel
\cite{Hsieh2008} and limited number of iterations ($100$ times the number of samples).
 The regularization hyperparameter $C$ of the SVM was optimized 
using $5$ fold cross validation with two repetitions and with the values 
$[10^{0},10^{-1},\ldots, 10^{-6}]$. 
To account for the class imbalance a class weight of $2$ was used inside the SVM.
Finally a threshold optimization was performed to adapt the decision boundary 
to the chosen performance metric, i.e., the balanced accuracy. 

\section{Temporal Distortions P300}
\label{sup:P300time}

\begin{figure}[!t]
  \begin{center}
\includegraphics[width=.49\linewidth]{P300_interSession_dataset_timeshift.pdf}
\includegraphics[width=.49\linewidth]{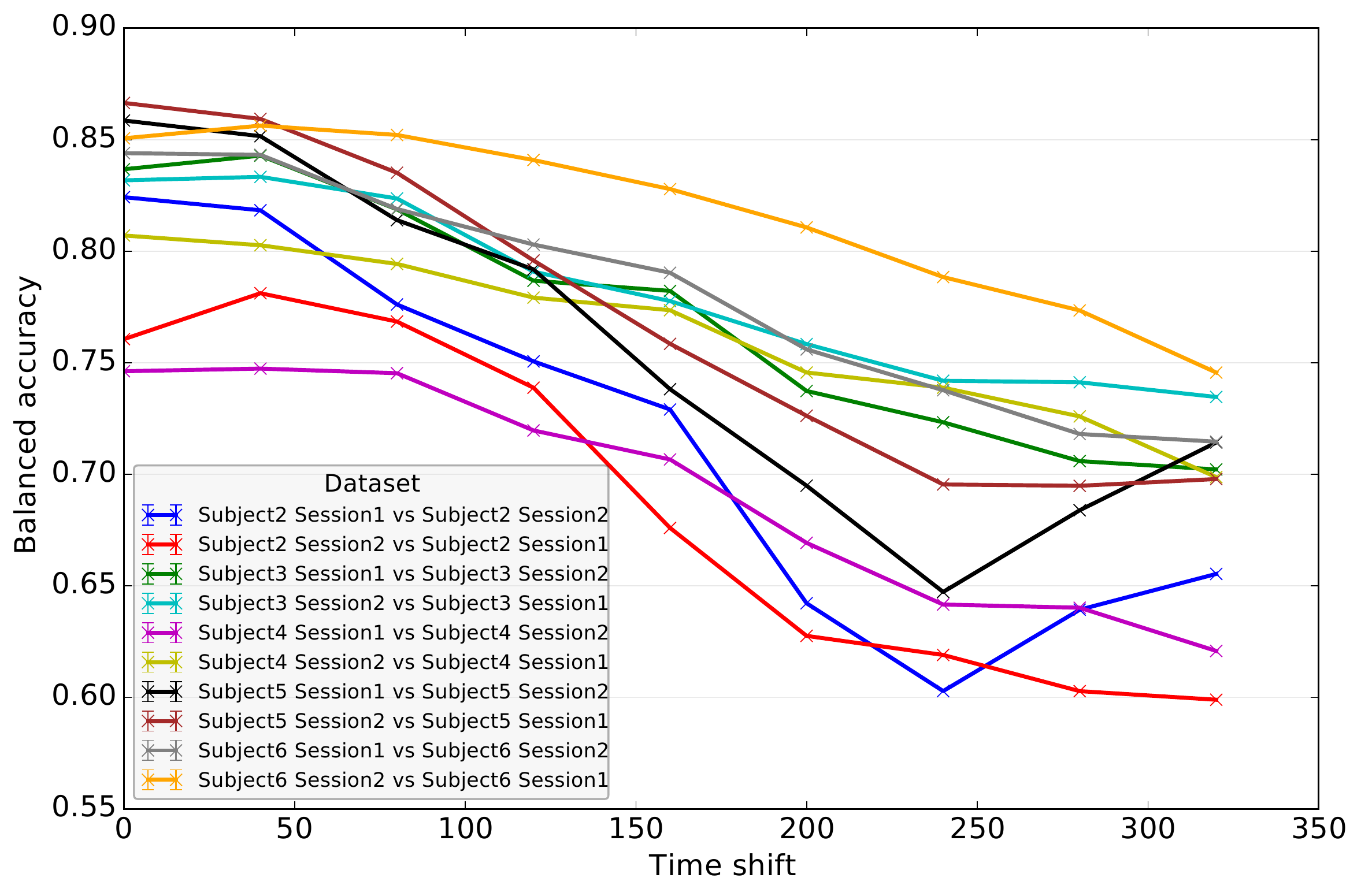}
  \end{center}
  \caption{Data augmentation through time shift in milliseconds in the setting
  of transfer between sessions. [left] Augmentation with one time shift additionally
  to the original data ($0$). [right] Augmentation with two time shifts: $\pm$ \emph{time shift}
  additionally to the original data.}
  \label{sup:time_shift}
\end{figure}

The results for the temporal are depicted in \Reffig{sup:time_shift}.
Here, zero was not the best choice for the offset
which is reasonable due to possible changes of the P300 offset
between experiments.
Furthermore, a change in the sign of the optimal time shifts was
observed when comparing two sessions 
which strengthens the impression that this a systematic effect.
On average, performance decreased when the original zero offset was not used.
But for the data augmentation, adding data with $40$ and $-40$ ms shift did not 
change performance on average but using larger offsets decreased performance.
Hence, this approach does not add temporal robustness in our case 
(P300 detection) but at least increases the number of samples.

\section{Interpolation Strategy}
\label{sup:inter_stra}

\begin{figure}[!t]
  \begin{center}
\includegraphics[width=.49\textwidth]{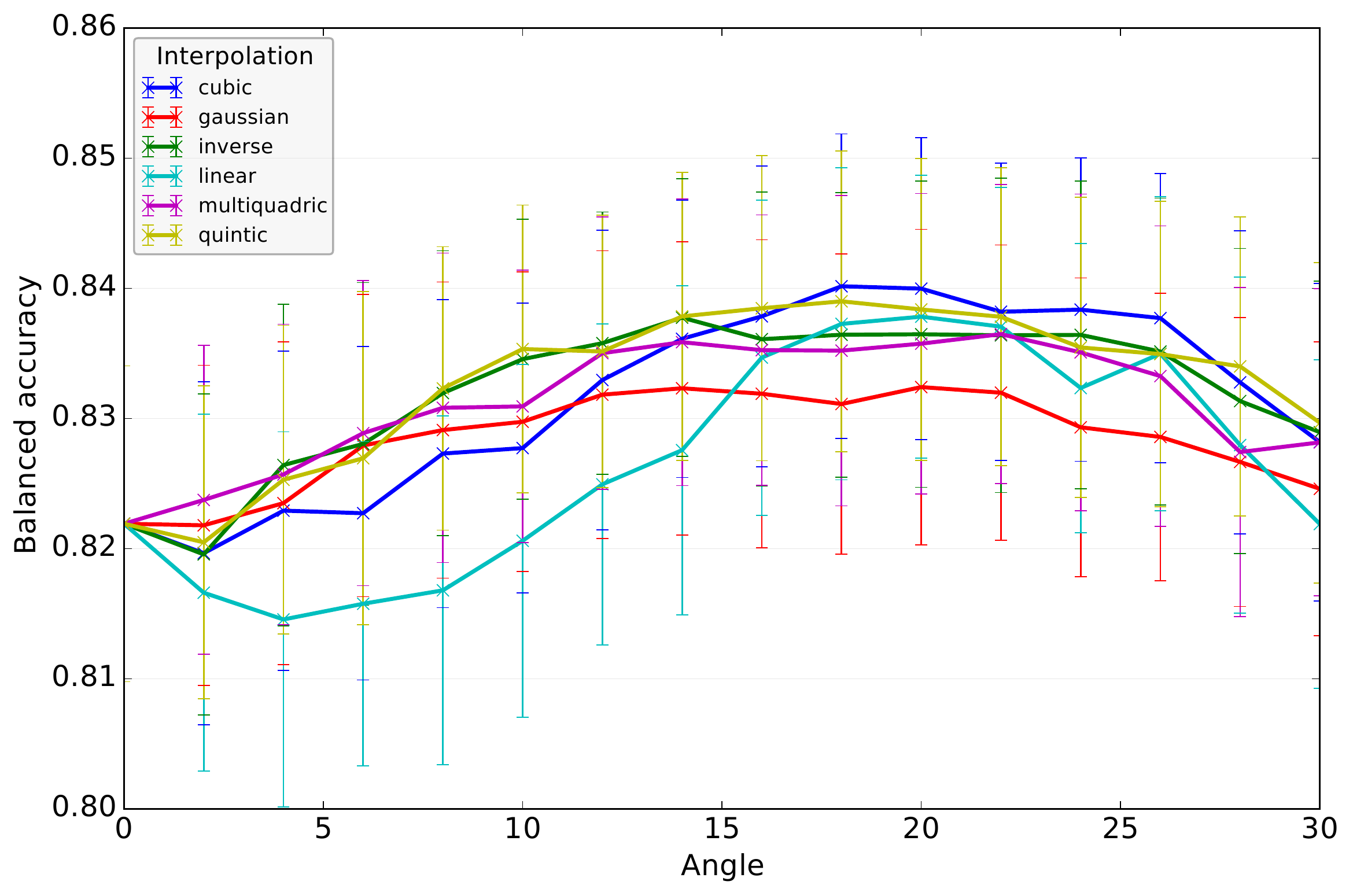} 
\includegraphics[width=.49\textwidth]{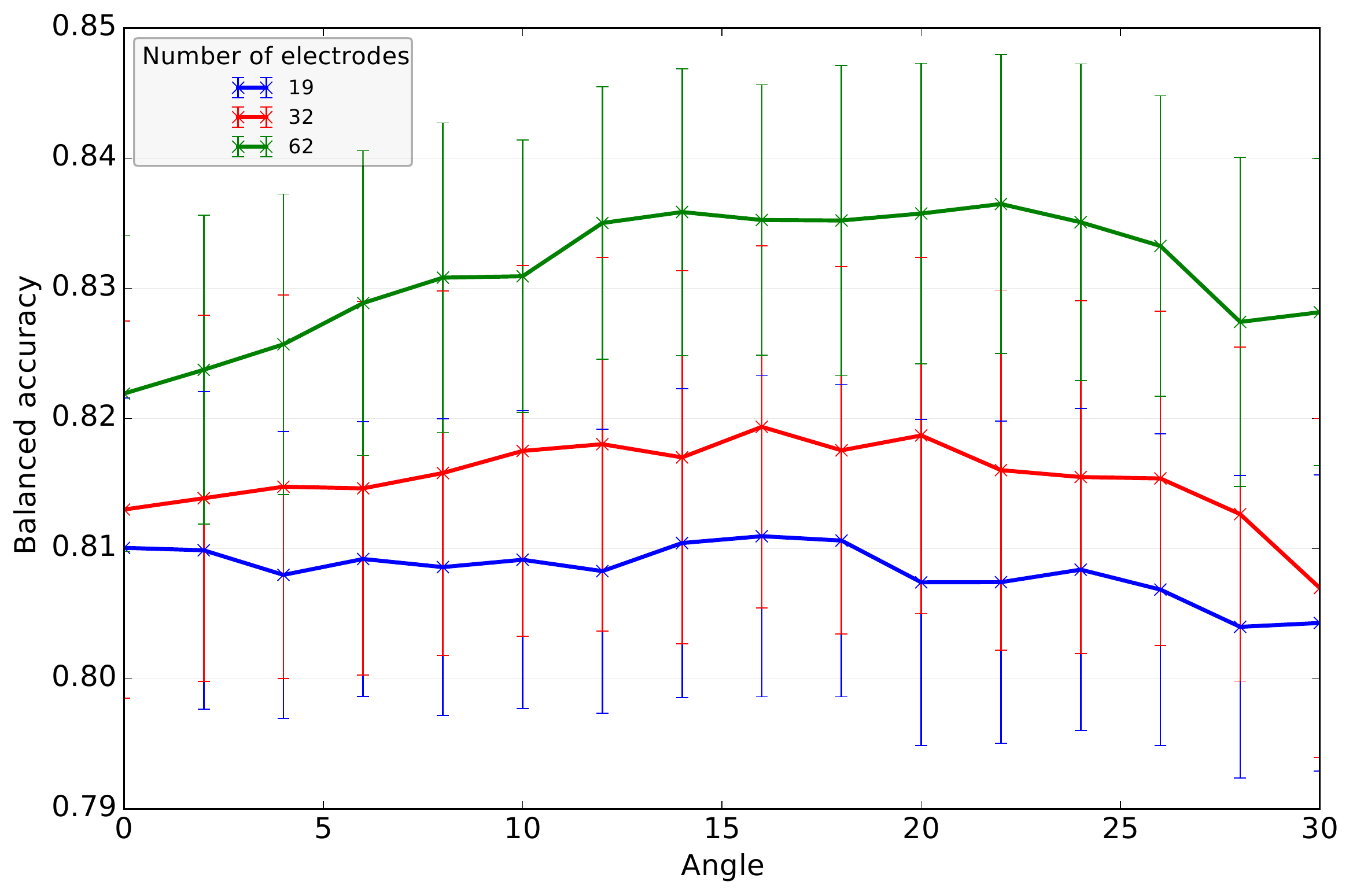}
  \end{center}
\caption{Comparison of interpolation strategies (left) and
different cap configurations containing 
a different number of electrodes (right).}
\label{sup:interpol}
\end{figure}

For the interpolation, different strategies could be used in SciPy.
In a preliminary analysis on synthetic data, we found out that
the \verb|scipy.interpolate.Rbf| library fits best to our 
research issue.
Due to the 3-dimensional positioning of the electrodes,
the interpolation is not straightforward.
Different parameters can be used for an internal modeling function.
So far, we used the default strategy (``multiquadric'').
As shown for \Reffig{sup:interpol} [left], 
the default method
performed quite well but slightly better results 
might be achieved with another interpolation function
like ``quintic'' or ``cubic''.

Further, 
we evaluated whether the data augmentation/interpolation is possible 
for smaller electrode constellations 
($32$ and $19$ electrodes according to the $10$-$20$ system).
This evaluation shows 
that the data augmentation did not reduce 
the performance for smaller electrode constellation,
but there was also no relevant improvement between angles
(see \Reffig{sup:interpol} [right]).
This indicates that the interpolation is not good enough if the electrodes
are positioned too sparse.

\section{Comparison of Rotation Axes for Inter-Session Evaluation on P300 Data}
\label{sup:axex_inter}

\begin{figure}[!t]
  \begin{center}
\includegraphics[width=.49\textwidth]{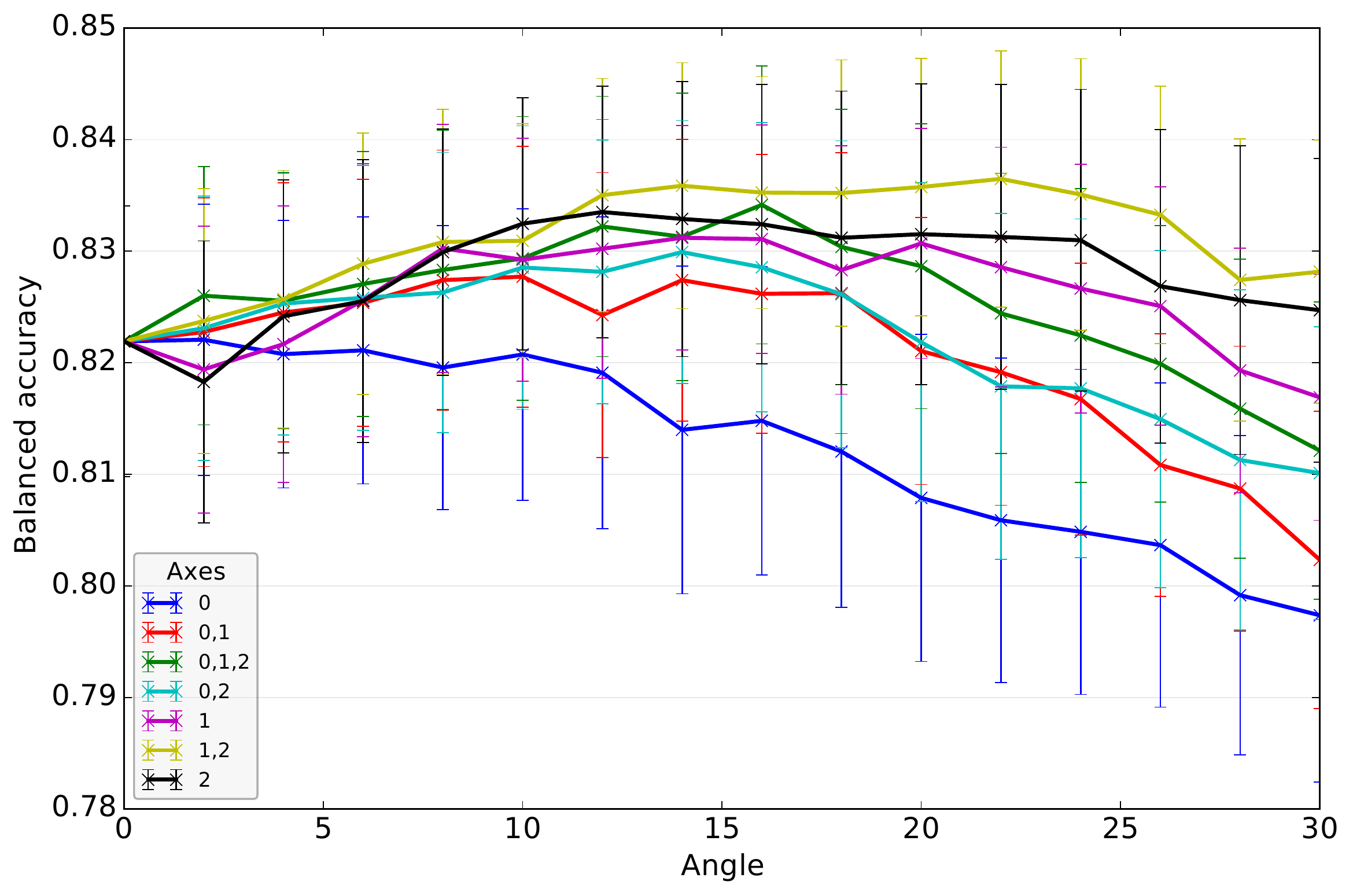} 
\includegraphics[width=.49\textwidth]{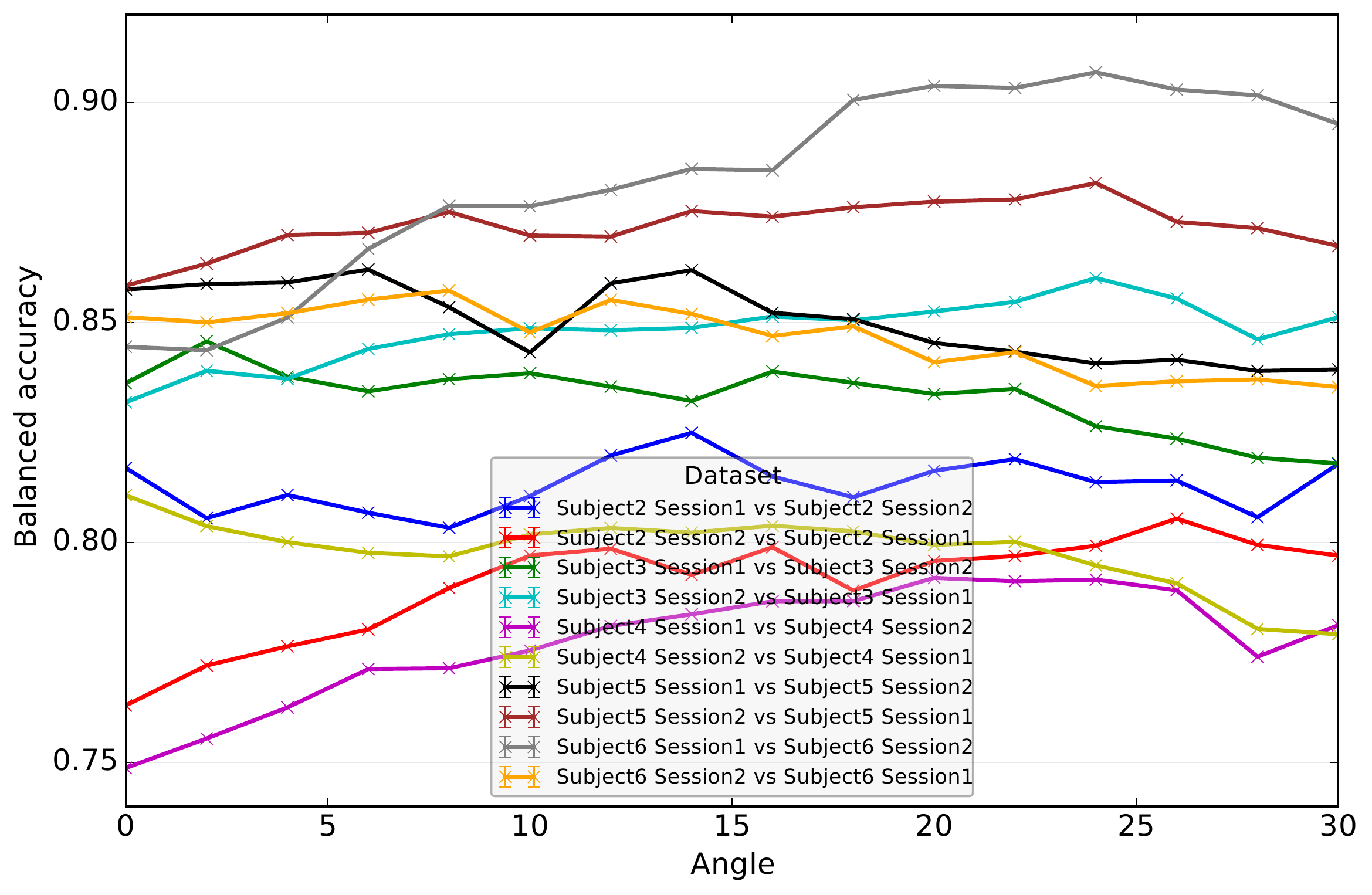}
  \end{center}
\caption{Data augmentation with a rotation 
around the different axes: x~(0), y~(1), z~(2) and a rotation
angle of \emph{+angle} \textbf{and} \emph{-angle} (combined).
For the left image, the augmented data of the different rotation axes is
combined for training.
The right picture shows the data augmentation with
rotation around the z- and the y-axis (corresponds to ``1,2'')
for each session transfer. 
}
\label{sup:directions_subject}
\end{figure}

Figure~\ref{sup:directions_subject} depicts the result for the comparison
on the inter-session evaluation on P300 data for the comparison different axes
as well as the individual results for the case of rotating around
z- and the y-axis. 

\section{Statistical evaluation}
\label{sup:stat}
\emph{In regards to Section~\textbf{\ref{s:dim_changes} Data Reduction and Data Dimensionality Increase:}}
For comparison between different number of filters and rotation angles (\Reffig{f:red} [left]),
we performed a two-way ANOVA with \emph{number of filters} and \emph{rotation angle} as within-subjects factors.
For multiple comparisons, Bonferroni-Holm was applied.
Rotation angle has an effect on the classification performance 
[$F_{15, 135} = 2.53, p < 0.031$,], 
but the number of filters does not [$F_{3, 27} = 0.96, p = n.s.$].
We observe the performance decrease for small rotation angles ($2\si{\degree}$),
but this performance reduction recovers when increasing rotation angles (between $14\si{\degree}$ and $20\si{\degree}$).
This pattern is obviously revealed for large data dimension (filter number of $64$)
[$2\si{\degree}$ vs. between $14\si{\degree}$ and $20\si{\degree}$: $p< 0.05$].
 
For comparison between training sizes and rotation angels (\Reffig{f:red} [right]),
we performed a two-way ANOVA with \emph{training size} and \emph{rotation angle} as within-subjects factors.
For multiple comparisons, Bonferroni-Holm was applied.
Both training size and rotation angels have an effect on the classification performance
[effect of training size: $F_{3, 27} = 66.99, p < 0.001$, effect of rotation angel: $F_{15, 135} = 3.52, p< 0.001$].
We found no interaction effect between both factors [$F_{45, 405} = 0.41, p = n.s.$].
Especially, performance increase through rotation is observed between $18\si{\degree}$ and $20\si{\degree}$
[no rotation ($0\si{\degree}$) vs. between $18\si{\degree}$ and $20\si{\degree}$: $p < 0.05$].
This kind of performance improvement is particularly revealed for the training size of $20\%$, i.e., 
a significant effect of rotation angle is only revealed when the training size is small.
A statistic with significant effects with larger training size
is provided for the subject transfer setting (Section~\ref{s:trans}).
For the transfer between P300 recording sessions,
the \emph{sample size} was only $10$ in contrast to $80$ and $56$
for the other comparisons.

\end{document}